\pdfoutput=1

\documentclass[11pt]{article}

\usepackage[final]{acl}

\usepackage{times}
\usepackage{latexsym}

\usepackage[T1]{fontenc}

\usepackage[utf8]{inputenc}

\usepackage{microtype}

\usepackage{inconsolata}
\usepackage{enumitem}
\usepackage{graphicx}
\usepackage{subcaption}
\usepackage{booktabs}
\usepackage{array}
\usepackage{multirow}
\usepackage[normalem]{ulem}
\usepackage{amsmath}
\usepackage[most]{tcolorbox}
\tcbuselibrary{listings,breakable}

\title{
\textit{Who's Asking?} Simulating Role-Based Questions\\for Conversational AI Evaluation}

\author{Navreet Kaur$^{1}$ \quad
Hoda Ayad$^{1}$\thanks{Equal contribution.} \quad
Hayoung Jung$^{2}$\footnotemark[1]\thanks{Work done while at the University of Washington.}\\
\textbf{Shravika Mittal}$^{3}$ \quad
\textbf{Munmun De Choudhury}$^{3}$ \quad
\textbf{Tanushree Mitra}$^{1}$\\
$^{1}$University of Washington \quad
$^{2}$Princeton University \quad
$^{3}$Georgia Institute of Technology\\
\texttt{kanavr@uw.edu}, \ \texttt{tmitra@uw.edu}
}

\definecolor{ptcolor}{HTML}{54A9E6}
\definecolor{cgcolor}{HTML}{F3856C}
\definecolor{prcolor}{HTML}{B484CF}

\newtcolorbox{promptbox}{
  breakable,
  enhanced,
  colback=gray!10,
  colframe=black!50,
  boxrule=0.3pt,
  arc=2pt,
  outer arc=2pt,
  left=0.2em,right=0.2em,top=0.1em,bottom=0.1em,
  listing only
}

\lstdefinestyle{promptstyle}{
  basicstyle=\ttfamily\footnotesize,
  breaklines=true,
  breakatwhitespace=false,
  columns=fullflexible,
  keepspaces=true,
  showstringspaces=false,
  keywordstyle=\color{blue}\bfseries,  
}

\newcommand{\methodname}{\textsc{CoRUS}~}

\begin{document}
\maketitle

\begin{abstract}
Language model users 
often embed personal and social context in their questions. 
The
asker's \textit{role}---implicit in
how the question is 
framed---creates specific needs for 
an appropriate response. However, most evaluations, while capturing the model's capability to respond, %
often 
ignore
\textit{who} is asking.  
This gap is especially critical in stigmatized domains such as opioid use disorder (OUD), 
where accounting for users' contexts is essential to provide accessible, stigma-free responses.
We propose \textbf{\methodname} (\textbf{CO}mmunity-driven \textbf{R}oles for \textbf{U}ser-centric Question \textbf{S}imulation), a framework for simulating role-based questions.
Drawing on role theory 
and 
posts from an online OUD recovery community (\texttt{r/OpiatesRecovery}),
we first build a taxonomy of asker roles---patients, caregivers, practitioners. Next, we use it to simulate $15{,}321$ 
questions that embed each role's \textit{goals}, \textit{behaviors}, and \textit{experiences}. Our evaluations show that these questions are both highly believable and 
comparable to real-world data.
When used to evaluate five LLMs, for the same question but differing roles, we find systematic differences: vulnerable roles, such as patients and caregivers, elicit more supportive responses ($+17\%$) and reduced knowledge content ($-19\%$) in comparison to practitioners. %
Our work demonstrates how implicitly signaling a user's role shapes model responses, and provides a methodology 
for 
role-informed 
evaluation of conversational AI.\footnote{We will release the datasets and code upon publication.} 
\end{abstract}

\section{Introduction}
\begin{figure*}[t]
    \centering
    \includegraphics[width=2.05\columnwidth]{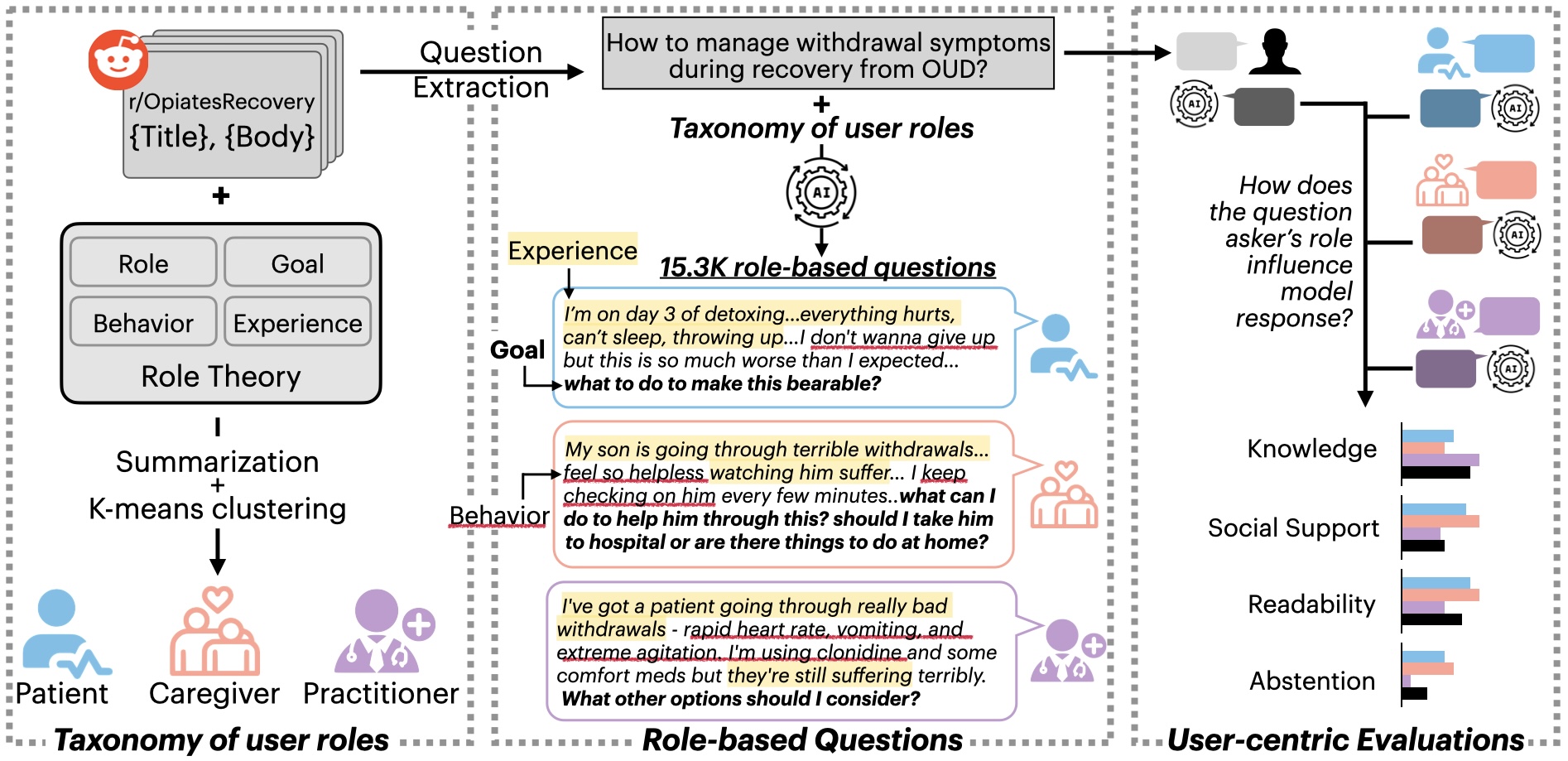}
    \caption{
    \textbf{\methodname framework.} Users' framing of a question often reflects the social context in which it is asked, yet most evaluations ignore this.
    We propose \methodname to (1) derive a taxonomy of askers' roles, grounded in role theory and posts from an online recovery community (\S\ref{sec:taxonomy}), (2) simulate role-based questions embedded with social context (\S\ref{sec:simulation}),  and (3) evaluate models to assess how role-based context influences response (\S\ref{sec:method-audit}). 
    }
    \label{fig:framework}
\end{figure*}

\begin{quote}
\small
\textbf{User 1 (patient):} I am on day 3 of detoxing from opiates. I can't sleep and keep throwing up. I don't wanna give up. How to make this bearable?

\textbf{User 2 (caregiver):} My son is withdrawing from opiates. I feel so helpless watching him suffer. What can I do to help him through this?
\end{quote}

In the examples above, both users pose the same question---``How to manage withdrawal symptoms?''---yet from different perspectives: first is from a \textbf{patient} recovering from withdrawal, while the second is from a \textbf{caregiver} caring for a loved one.  
These perspectives reflect users' \textbf{roles} (\S\ref{sec:background}), 
which capture the \textit{goals}, \textit{behaviors}, and \textit{experiences} that shape how they frame questions (\S\ref{sec:taxonomy}).
Users often 
share such personal and social
context 
with large language model (LLM) powered conversational AI systems \cite{mireshghallah2024trust}
to receive 
tailored advice 
\cite{zhang2024s}.

Yet, most evaluations of such systems
strip away this context \cite{malaviya2024contextualized}, 
ignoring
the social framing that shapes how people actually ask (\S\ref{sec:background}).
This limitation is especially consequential
in stigmatized domains 
like opioid use disorder (OUD)---a leading cause of death in the U.S. 
\cite{NIDA2023OverdoseDeathRates}---where stigma
and 
mistrust limit access to 
care 
\cite{cernasev2021systematic, woo2017don}. 
As a result, 
people increasingly turn to online communities \cite{balsamo2023pursuit}, and now to LLMs \cite{choy2024can} for treatment planning, and even seeking hope \cite{anthropic2025affective}. Thus the same question posed to an LLM may %
carry different expectations depending on \emph{who} is asking (as seen in our opening examples). %
While LLMs invite
\textit{role}-specific disclosure
due to
perceived anonymity and judgment-free interaction, 
we still lack systematic understanding of how an asker's \textit{role} and their social context---implicitly signaled in their queries---influence responses in stigmatized domains. %
Simply put, current LLM evaluations are role-agnostic and fail to capture the nuances of \emph{who's asking}.

In this work, 
we propose 
\textbf{\textsc{CoRUS}}---\textbf{CO}mmunity-driven \textbf{R}oles for \textbf{U}ser-centric Question \textbf{S}imulation---a framework 
that embeds role-based context implicitly into a given question
for LLM evaluation,
drawing on how people share {personal and social context} in their queries on online communities (Figure \ref{fig:framework}).
By applying this framework to OUD recovery, 
we investigate three main research questions.

First, 
we investigate \emph{who} brings their questions into an OUD recovery community
(\S\ref{sec:taxonomy}).
Drawing on role theory \cite{biddle1979role, turner1990role},
which conceptualizes roles as socially structured expectations shaping how individuals behave and how others respond,
we
derive
\textbf{a taxonomy of roles}
using
posts
on \texttt{r/OpiatesRecovery}---an online community dedicated to OUD recovery discussions.
We identify three primary roles defined by their 
\textit{goals}, \textit{behaviors}, and \textit{experiences} (see Table \ref{tab:taxonomy-small}).

Second,  
we study whether \textit{believable} questions---those that are natural, realistic, or close proxies of questions asked by humans and aligned with the asker's social context
\cite{park2023generative, zhou2024sotopia}---can
be simulated 
for
different roles (\S\ref{sec:simulation}).
Questions must be believable 
to mimic how people actually ask; otherwise, they remain detached from real use and lose ecological validity \cite{liu2025understanding}.
We simulate role-specific variants of questions extracted from 
\texttt{r/OpiatesRecovery},
by leveraging our taxonomy to embed each role's \textit{goals}, \textit{behaviors}, and \textit{experiences}.
Our evaluations show that
$90\%$ of these questions are 
as believable as
real-user queries from WildChat \cite{zhao2024wildchat} and Reddit,
and $99\%$ faithfully reflect each role's perspective
(\S\ref{sec:sim-eval-results}).
We contribute 
$15{,}321$ role-specific questions, 
a \textbf{large-scale resource for evaluating LLMs 
beyond context-free questions.}

Third,
we evaluate five user-facing LLMs 
to examine how
role-based contexts 
change responses 
relative to role-agnostic queries
(\S\ref{sec:method-audit}).
Unlike existing evaluations that ignore who is asking, \methodname reveals systematic shifts
(\S\ref{sec:results-audit}): 
patient and caregiver contexts reduce knowledge ($\geq15\%$) 
but increase support ($\geq15\%$)
and readability ($\geq24\%$); 
practitioner context 
leaves knowledge unchanged, 
and decreases support ($9\%$) and readability ($15\%$).
\textbf{While models 
provide supportive
responses,
they are less informative 
for those most in need}
\citep{eslami2021information, mardani2023challenges}, %
posing a challenge to the usefulness of LLMs in sensitive domains.

Overall, we show that
role-agnostic 
evaluations, which ignore the asker's role,
miss systematic differences in how LLMs respond to the same question across users.
Overlooking such role-based variation risks 
designing systems that 
fail to meet distinct user needs.
While we focus on OUD recovery, 
\methodname is 
extendable to other domains, and
we hope 
it
inspires more 
role-aware evaluation of conversational AI,
where social context 
shapes human-AI interaction \cite{liao_rethinking_2025}.

\section{Why Role-based Evaluations?}
\label{sec:background}

\textit{What} users share, and \textit{how} they phrase questions, 
implicitly represent their needs, expertise, and lived experiences \cite{chan2010decomposing, yang2019seekers, saxena2022users, tseng-etal-2024-two} i.e. their \textit{role}.
Role theory 
conceptualizes 
a \textit{role} as the socially structured expectations that shape
how an individual behaves and how others respond
\cite{biddle1979role},
which 
varies with circumstances \cite{turner1990role}.
Asking a question is thus a performance of role, 
enacted
through framing and disclosure 
\cite{goffman1949presentation}.
Questions are therefore not role-agnostic requests, but social acts that encode the asker's role and expectations.%

Such role-based questions
are especially critical in stigmatized, high-stakes domains, where users seek not just information but also 
support. 
In OUD recovery, individuals turn to online communities \cite{balsamo2023pursuit}, and increasingly to LLMs for guidance \cite{choy2024can}, as 
stigma limits access to care \cite{cdc_stigma_2023}.
Crucially, what counts as an appropriate response to 
the same question
depends on who asks it---a distinction that current role-agnostic evaluations fail to capture (see Figure \ref{fig:framework}).
For instance, 
a generic response to 
\textit{``How to manage withdrawal symptoms?''}
that suggests seeking medication, tapering opioids, and warns about relapse may seem reasonable; however, it can fail to comfort a distressed patient, be impractical for a caregiver, or lack rigor for a practitioner.
Current role-agnostic evaluations fail to identify such nuanced failures,
a gap we address in this work.

\section{Taxonomy of Information-seeking Roles}
\label{sec:taxonomy}
\begin{table*}[t]
\small
\centering
\resizebox{\linewidth}{!}{\begin{tabular}{@{}m{0.1\columnwidth} m{0.55\columnwidth} m{0.55\columnwidth} m{0.55\columnwidth}@{}}
\toprule
\multicolumn{1}{c}{\textbf{Role}} 
& \multicolumn{1}{c}{\textbf{\colorbox{ptcolor!50}{Patient}}}  
& \multicolumn{1}{c}{\textbf{\colorbox{cgcolor!50}{Caregiver}}} 
& \multicolumn{1}{c}{\textbf{\colorbox{prcolor!50}{Practitioner}}}\\
\midrule
\multicolumn{1}{c}{\textbf{Description}}
& Individuals in different stages of OUD recovery 
& Family, friends, or peers providing informal support during recovery
& Professionals treating patients e.g., clinicians, nurses, counselors \\
\midrule
\multicolumn{1}{c}{\textbf{Goal}}
& Seek specific guidance for themselves and share experiences with withdrawal, treatment, coping
& Seek advice on treatment logistics for patients, and discuss the personal toll of caregiving 
& Seek \& share clinical, therapeutic, and harm-reduction information; learn about patient experiences\\
\midrule
\multicolumn{1}{c}{\textbf{Behavior}}
& Use informal conversational tone with emotional, vulnerable health-related self-disclosure 
& Use informal, emotive language expressing concerns and responsibilities of caregiving
&  Use direct, authoritative tone with a mix of conversational and technical language, hinting expertise \\
\midrule
\multicolumn{1}{c}{\textbf{Experience}} 
& Lived experience with active substance use or early recovery 
& Witness loved one's struggles and manage its practical impacts                                                         & Working with diverse patients and using different treatments\\
\midrule
\multicolumn{1}{c}{\textbf{Example Post}} 
& \textit{I'm 1 year sober from fent after 13 years IV use, but still have strong urges thinking I can handle it. How long will cravings last?}
& \textit{My partner is detoxing at my place, is angry and hostile. I know it's a withdrawal symptom, but what if it's beyond 1-2 weeks?}
& \textit{After 30 yrs treating addiction, I think `relapse is part of recovery' is dangerous. Can sobriety last with commitment, support?}
\\ \bottomrule
\end{tabular}}
\caption{\textbf{Taxonomy of information-seeking roles.} Roles are characterized by their \textit{goals}, \textit{behaviors} and \textit{experiences}, derived from \texttt{r/OpiatesRecovery} posts, with each illustrated by a paraphrased example post. 
}
\label{tab:taxonomy-small}
\end{table*}

We develop a taxonomy of 
information-seeking 
roles by combining a top-down framework grounded 
in role theory \cite{biddle1979role, turner1990role, yang2019seekers}
(\S\ref{sec:social-role-framework}) with a large-scale, bottom-up analysis of Reddit posts (\S\ref{sec:taxonomy-construction}).

\subsection{
Operationalizing \textit{Role}}
\label{sec:social-role-framework}
Informed by role theory \cite{biddle1979role, turner1990role}, we adapt 
\citet{yang2019seekers}'s 
framework to 
characterize information-seeking roles in OUD recovery through the following facets: (i) \textbf{Goals}: intents such as seeking information or emotional support, or disclosing struggles; (ii) \textbf{Behaviors}: observable linguistic features, including emotional expressions, mentions of medical terms, or references to external resources; and (iii) \textbf{Experiences}: descriptions of lived experiences, reflecting comfort with disclosing sensitive information.

\subsection{Taxonomy Construction}
\label{sec:taxonomy-construction}
Recent work shows that LLM-based summarization and clustering pipelines can support large-scale data analysis 
\cite{tamkin2024clio, lam2024concept, wan2024tnt, rao2024quallm}. We build on this methodology
to construct a taxonomy of information-seeking roles in OUD recovery community, 
grounding the analysis in role-specific facets i.e. \textit{goal}, \textit{behavior} and \textit{experience} (\S\ref{sec:social-role-framework}). 
Our method blends inductive coding with deductive guidance from prior work \cite{yang2019seekers, biddle1979role, turner1990role}:
while LLMs surface preliminary codes, human judgments remain central in 
refining the taxonomy. Below we outline the key steps for constructing the taxonomy.

\textbf{(1) Summarization.}
We collect $10{,}017$ posts between January 1, 2023, and December 31, 2024, from \texttt{r/OpiatesRecovery}, a Reddit community ``dedicated to helping each other stop and stay stopped'' from opioid use.
Following \citet{tamkin2024clio}, we scale qualitative analysis by using Claude 3 Haiku (\texttt{claude-3-haiku-20240307}) to generate facet-based summaries (\textit{goal}, \textit{behavior}, \textit{experience}) per post (examples in Table \ref{tab:facet-summary-examples}), mirroring the initial coding phase of thematic analysis \cite{clarke2017thematic}.
\textbf{(2) Clustering.}
We embed these summaries 
using a sentence transformer (\texttt{all-mpnet-base-v2}) \cite{reimers2019sentence, NEURIPS2020_c3a690be}, 
and cluster them with k-means, selecting four clusters based on UMAP visualization \cite{mcinnes2018umap} and Silhouette analysis \cite{rousseeuw1987silhouettes} (see Appendix \ref{sec:appendix-taxonomy} for details). 
We assign preliminary cluster descriptions using Claude 3.7 Sonnet (\texttt{claude-3-7-sonnet-20250219}) and refine each by manually reviewing $40$ posts per cluster (see Appendix \ref{sec:appendix-taxonomy} for details). 
\textbf{(3) Human Validation.}
Two expert annotators with prior experience in health NLP research validate both generated facet-based summaries and cluster labels, showing high inter-rater agreement and close alignment with model outputs (Cohen’s $\kappa~0.71{-}0.93$, accuracy $0.70{-}0.88$; see Tables \ref{tab:irr-taxonomy-summary}–\ref{tab:irr-taxonomy-cluster}). Further details are in Appendix \ref{sec:appendix-taxonomy}.

\paragraph{Taxonomy Structure.}
Through this process, 
we construct a taxonomy of three information-seeking roles in OUD recovery community: 
\textbf{Patient}, \textbf{Caregiver}, and \textbf{Practitioner}, each characterized by specific goals, behaviors and experiences (Table \ref{tab:taxonomy-small}).\footnote{We use the term \textit{information-seeking} broadly to include requests for information, support, or guidance.}
We also identify a {Community Participant} role, but 
focus on the former three---most relevant to conversational AI use \cite{NBERw34255}---and discuss the rest in Appendix \ref{sec:appendix-taxonomy} (Table \ref{tab:taxonomy}).

\section{Simulating Role-based Questions}
\label{sec:simulation}

Datasets that capture how different user roles frame questions do not exist (\S\ref{sec:background}), and simply
prompting LLMs to generate synthetic data often drifts from real-world style and distribution \cite{veselovsky2023generating}.
Moreover, explicitly specifying demographic or identity attributes in prompts risks stereotyping \cite{cheng2023compost, dammu-etal-2024-uncultured}.
To address these challenges,
we embed roles \textit{implicitly} in queries.
We first extract information-seeking questions from \texttt{r/OpiatesRecovery}, 
and then use our taxonomy 
of information-seeking roles 
(\S\ref{sec:taxonomy}) to generate role-framed variants.

\begin{table*}[t]
\centering
\small
\resizebox{\linewidth}{!}{\begin{tabular}{@{}m{0.9\columnwidth} m{1.1\columnwidth}} 
\toprule
{\textbf{Metric}} 
& \textbf{Definition} \\ \midrule
{Human Likeness \newline \cite{li2017adversarial, mir2019evaluating}}
& Whether a question is indistinguishable from a human-written one; this captures fluency and naturalness. \\ \cmidrule{1-2}
{Context Plausibility \newline \cite{gunther2021assessing, park2024valuescope}}
& Whether the context embedded in a question is plausible within the discourse of a relevant online community; this captures whether simulated questions reflect authentic narratives. \\ \cmidrule{1-2}
{Interaction Plausibility  \cite{balog2024usersimulationevaluatinginformation}}  
& Whether the question is plausibly phrased for a one-on-one conversation in a human-AI interaction setting; this reflects the realism of simulated questions within the setting in which they are posed. \\ \cmidrule{1-2}
{Role Faithfulness \cite{peng2024quantifying, ji2025enhancing}}  
& Whether the question reflects the perspective of the target role.  \\ \cmidrule{1-2}
{Content Preservation \newline \cite{fu2018style, mir2019evaluating, park2024valuescope}}   
& Whether the simulated question preserves the content of the role-agnostic information-seeking question. \\ 
\bottomrule
\end{tabular}}
\caption{Metrics for evaluating the \textit{believability} of simulated role-based questions (\S\ref{sec:sim-eval-results}). 
}
\label{tab:sim-eval-metrics}
\end{table*}

\paragraph{Curating Information-Seeking Questions.}
Following \citet{kumar_compo_2024}, we 
filter $10{,}017$ posts to retain only English, non-deleted, text-only posts, excluding moderator posts.
We detect questions by checking if the title, or first or last two body sentences end with 
a question mark
($86\%$ precision by manual check).
We then prompt GPT-4o-mini 
to rewrite each as a role-agnostic question, 
removing any
personal details and 
role-revealing cues.
Two 
authors
validate $50$ samples, confirming that $94\%$ rewrites preserve the original question,
and that only $0.06\%$ questions have role leakage (Cohen’s $\kappa$ = $0.82$). GPT-4.1, validated against these human labels ($72{-}78\%$ accuracy), provides additional filtering, resulting in $5{,}107$ 
information-seeking questions (see Appendix \ref{sec:appendix-seed-question-eval} for details).

\paragraph{Role-Based Simulation Design.}
We simulate role-based questions by combining:
(i) \textit{information-seeking
questions} that capture topical diversity,
and 
(ii) \textit{taxonomy of information-seeking roles} (\S\ref{sec:taxonomy}), that encodes role-based attributes. 
For each seed question, we 
condition 
Claude-3.7-Sonnet 
on these attributes,
along with sampled
facet-based summaries 
(Table \ref{tab:facet-summary-examples}),  
enriching simulated questions with behavioral and experiential nuances (details are in Appendix \S\ref{sec:appendix-sim-method}).
Applying \methodname to $5{,}107$ 
information-seeking questions across three roles---Patients, Caregivers, and Practitioners---yields 
$15,321$ role-based questions (examples in Table \ref{tab:query-examples}).
Next, we demonstrate the usefulness of the taxonomy in simulating these realistic role-based queries (\S\ref{sec:sim-eval-results}).

\subsection{Evaluating Simulated Questions}
\label{sec:sim-eval-results}

\paragraph{Metrics.}
\label{sec:sim-eval-metrics}
To ensure ecological validity \cite{liu2025understanding}, simulated questions should mimic how people actually ask. 
We therefore evaluate whether \methodname generates \textit{believable} role-based questions.
Prior work defines \textit{believability} as behavior perceived as natural, realistic and role-aligned  \cite{zhou2024sotopia, park2023generative}.
We operationalize this 
through five metrics grounded in 
style transfer, role-playing, and user-modeling literature, outlined in Table \ref{tab:sim-eval-metrics}. Examples of questions that satisfy or do not satisfy these metrics are in Table \ref{tab:sim-eval-metrics-examples}.

\paragraph{Evaluation Setup.} 
We use the following evaluation setup: 
(1) \textbf{Automatic}: GPT-4.1 judges \cite{zheng2023judging} all questions on all metrics (Table \ref{tab:sim-eval-metrics}) using binary labels; 
for content preservation, we collapse five rating levels to binary following \citet{briakou2021evaluating} (prompts are in Appendix \ref{sec:appendix-sim-auto-eval}).
(2) \textbf{Human}: We validate each GPT-4.1 judge with $7{,}365$ human annotations across $491$ queries (Table \ref{tab:sim-eval-distribution}).
Three annotators, recruited via Prolific, independently perform the same task as GPT-4.1 rater
(Gwet's AC1 $0.63{-}0.99$; GPT-4.1 accuracy $0.70{-}0.90$). Full details are in Appendix \ref{sec:appendix-sim-human-eval}.

We evaluate \textsc{CoRUS}-simulated questions against two sets  (Table \ref{tab:sim-eval-distribution}):
(i) \textbf{real-user queries} from \texttt{r/OpiatesRecovery} and 
Wildchat \cite{zhao2024wildchat}, 
as a reference to test believability,\footnote{Wildchat lacks OUD-specific queries, so we use health- and clinical-related questions from the dataset, identified using task and domain labels from \citet{mireshghallah2024trust}.}
and 
(ii) \textbf{prompting variants}---\textsc{R}, 
\textsc{RG}, 
\textsc{RGB}, 
and \textsc{RGBE} (\texttt{role+goal+behavior+experience})---as ablation baselines against the full \methodname pipeline that also incorporates behavior- and experience-based summaries for greater nuance.

\subsection{Results}

\begin{figure}[t]
\centering
\includegraphics[width=\columnwidth]{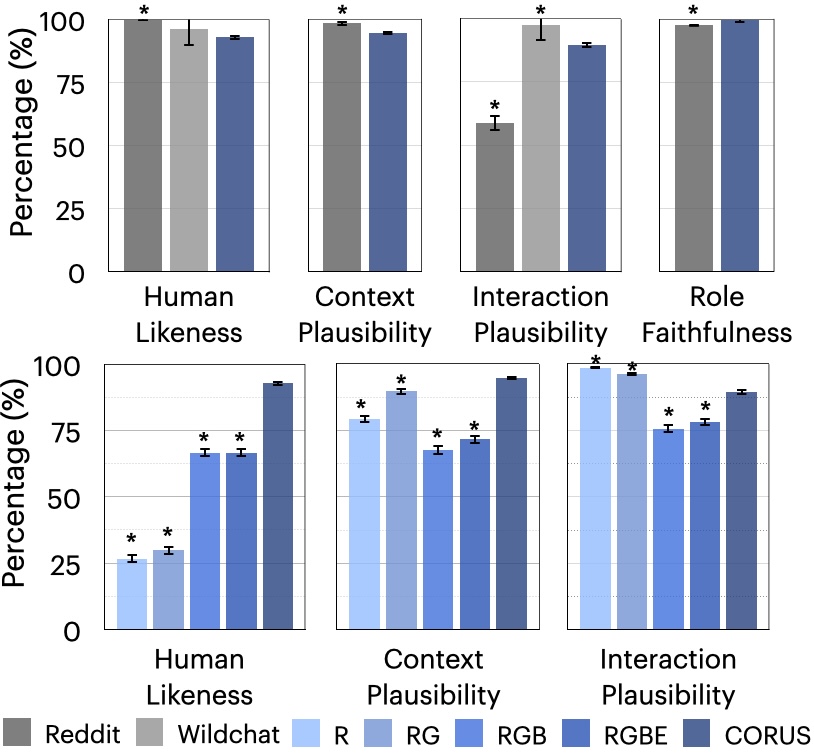}
\caption{\textbf{Evaluation of Simulated Role-based Questions.} Top: \methodname produces questions with \textit{believability} comparable to real-world ones. 
Bottom: \methodname produces significantly more human-like and contextually plausible queries than other prompting methods, while maintaining high interaction plausibility and other properties 
(Appendix \ref{sec:appendix-sim-eval}, Figure \ref{fig:appendix-corus-eval}). * denotes significant difference from \methodname (Chi-square test, $p<0.05$); error bars show $95\%$ CI.}
\label{fig:corus-eval} 
\end{figure}

\begin{figure*}[t]
\centering
\includegraphics[width=2.07\columnwidth]{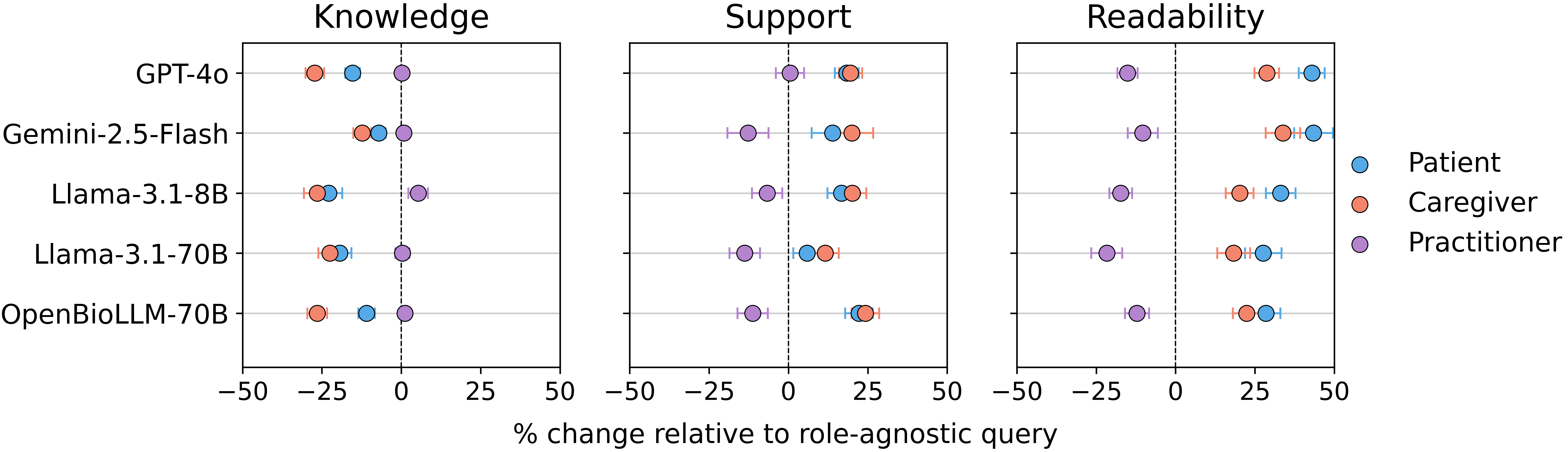} 
\caption{\textbf{Percentage change in Knowledge, Support and Readability scores of model responses when role-based context is added to the queries.} 
Across all models, 
patient 
and caregiver 
context reduces knowledge, but increases support 
and readability.
Practitioner context shows the inverse trend, with lower support, 
higher readability, 
and no significant change in knowledge. More details 
in Figures \ref{fig:llm-eval-context} and \ref{fig:box-plot-llm-eval}. Error bars represent 95\% CI. 
} 
\label{fig:response-scores-change}
\end{figure*}

\paragraph{Comparison with Real-User Queries.}
\textsc{CoRUS}-generated queries achieve believability comparable to real-user queries (Figure \ref{fig:corus-eval}, top). 
$93\%$ queries are judged \textit{human-like}, closely matching WildChat ($96\%$) and Reddit ($100\%$).
$94\%$ are judged \textit{contextually plausible} within \texttt{r/OpiatesRecovery}, similar to Reddit ($98\%$); WildChat is not comparable here, as it lacks OUD-specific questions.
For \textit{interaction plausibility}, $89\%$ are judged plausible as chatbot-directed questions, 
compared to  $59\%$ for Reddit and $97\%$ for WildChat.
The lower proportion for Reddit reflects that community posts are phrased for collective audiences rather than conversational AI systems.
\textit{Role faithfulness} holds in $99\%$ of cases ($99\%$ patient, $98\%$ caregiver, $100\%$ practitioner), comparable to Reddit ($98\%$).
\textit{Content preservation} requires paired inputs and is therefore not applicable in this comparison.
Overall, \methodname produces questions that are human-like, contextually plausible, directed towards a chatbot, and role-faithful---achieving believability on par with real-world queries.\footnote{While some are significantly different (Figure \ref{fig:corus-eval}, top), effect sizes are small to moderate ($0.2{-}0.4$), 
except for Reddit's \textit{interaction plausibility} ($0.7$) (details are in Appendix \ref{sec:appendix-results}).}

\paragraph{Comparison with Prompting Baselines.}
\methodname outperforms the prompting baselines (Figure~\ref{fig:corus-eval}, bottom).
$93\%$ of \methodname generated queries are judged \textit{human-like}, compared to $27\%-67\%$ for baselines. \textit{Contextual plausibility} also increases to $95\%$, compared to $68\%-90\%$. For \textit{interaction plausibility}, $89\%$ of \methodname queries are judged plausible, lower than \textsc{R} ($99\%$) and \textsc{RG} ($96\%$), but higher than more complex baselines ($76\%-78\%$). The higher proportions for \textsc{R} and \textsc{RG} reflect overly direct phrasing that looks plausible but lacks naturalness and community grounding.
In contrast, \methodname balances human-likeness, and contextual and interaction plausibility.
\textit{Role faithfulness} remains high across all methods ($\sim99\%$), but $18\%-72\%$ of baseline queries rely on explicit role mentions (such as ``I am a patient \ldots''), while \methodname achieves role faithfulness without such mentions (Table \ref{tab:explicit-role-mentions}). 
\textit{Content preservation} is also high, with $96\%$ of \methodname-generated questions retaining the intent of the source question while naturally adjusting framing and style.
Overall, \methodname produces more human-like and contextually plausible queries than baselines, while maintaining high interaction plausibility, role faithfulness, and content preservation.

\section{Application: Role-based Evaluations}
\label{sec:method-audit}

We demonstrate the utility of \methodname 
by using the simulated questions to evaluate
how an asker's role-based context influences model responses. 

\subsection{Models Evaluated}
We generate $15{,}321$ role-specific questions across 
patients, caregivers, and practitioners,
and use them to evaluate 
five 
LLMs deployed in user-facing applications:
GPT-4o \cite{hurst2024gpt}, 
Gemini-2.5-Flash \cite{deepmind2025gemini25flash},
Llama-3.1–8B, Llama-3.1-70B \cite{grattafiori2024llama}, and Llama3-OpenBioLLM-70B, a specialized medical LLM \cite{pal2024openbiollms}. Experimental details are in Appendix \ref{sec:appendix-exp-details}.

\subsection{Response Evaluation Metrics}
\label{sec:audit-metrics}
Grounded in users' \textit{goals} from our taxonomy (Table \ref{tab:taxonomy-small}), sociology and health communication literature, we evaluate responses along the following:
\begin{itemize}[leftmargin=*, itemsep=0.0em, topsep=0.0em]
    \item \textbf{Knowledge}  \cite{fiske2007universal, levin2004strength}: Presence of guidance-oriented content, indicating the amount of information conveyed. Since people seeking OUD recovery information often lack access to medical guidance, information provision is critical \cite{eslami2021information, basak2025information, cernasev2021systematic}. 
    \item \textbf{Support} \cite{baumeister2017need, fiske2007universal, vaux1988social}: Expressions of empathy, reassurance or emotional aid. 
    Supportive communication is especially important in OUD recovery, as stigma discourages 
    seeking information \cite{cernasev2021systematic}. 
    \item \textbf{Readability}: Ease with which responses can be understood. Low health literacy is correlated with worse health outcomes \cite{wolf2005health}, and inaccessible language can further exacerbate barriers to care in OUD contexts \cite{cernasev2021systematic}. We measure readability using the Flesch Reading Ease score \cite{flesch1948new}, normalized to lie in $0{-}1$.
\end{itemize}

We adopt the classifiers from \citet{choi2020ten} to measure knowledge and support (scored between $0$ and $1$), validating them for our setting through $160$ human annotations (Cohen’s $\kappa$ $0.7$, accuracy $0.9$; details are in Appendix \ref{sec:appendix-exp-details}).

\subsection{How does the asker's role-based context shape LLM responses?}
\label{sec:results-audit}

\begin{figure}[t]
\centering
\includegraphics[width=0.9\columnwidth]{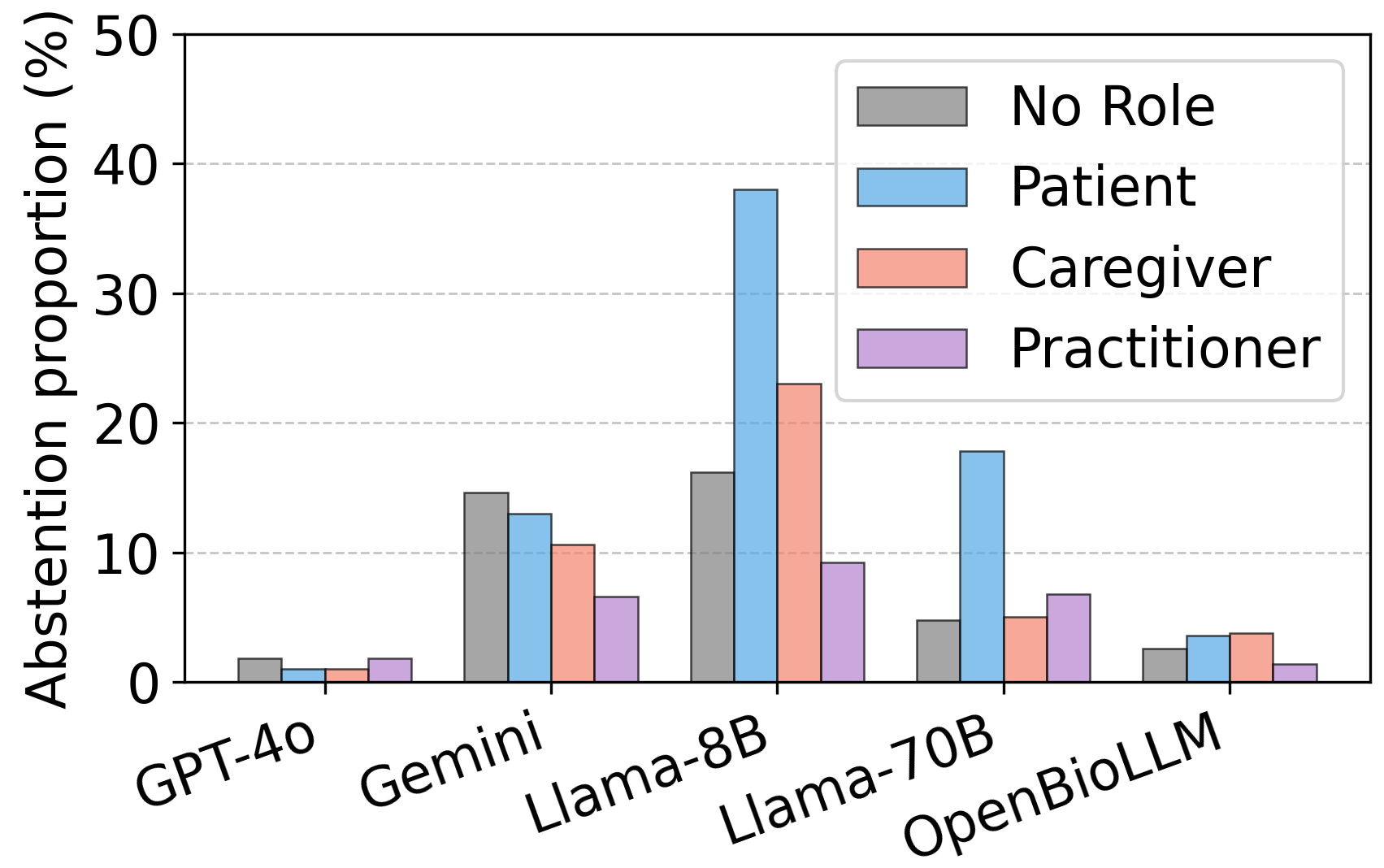} 
\caption{\textbf{Abstention.} Proportion of role-agnostic and role-based queries that models refuse to answer.}
\label{fig:abstention}
\end{figure}

We show how role-based context shifts responses 
across knowledge, support, and readability (\S\ref{sec:audit-metrics}), reporting percentage changes relative to role-agnostic queries (Figure \ref{fig:response-scores-change}). We also analyze cases where queries are refused (Figure \ref{fig:abstention}).

\paragraph{Patient.}
Embedding patient context consistently \textbf{reduces knowledge}, with an average drop of $15\%$ across models, largest in Llama-3.1-8B ($23\%$) and smallest in Gemini-2.5-Flash ($7\%$). In contrast, \textbf{support increases} by $15\%$ on average, from $6\%$ in Llama-3.1-70B to $22\%$ in OpenBioLLM-70B. \textbf{Readability improves} substantially, increasing from $27\%$ to $43\%$, highest in Gemini-2.5-Flash.

\paragraph{Caregiver.}
For caregiver context, models again \textbf{trade off knowledge for support and readability}.
Knowledge scores show the largest drop across roles ($23\%$ on average), with largest drop in OpenBioLLM-70B ($26\%$) and GPT-4o ($27\%$).
In contrast, support is highest for this role, with a $19\%$ increase on average, with largest increase in OpenBioLLM-70B ($25\%$).
Readability also improves by $24\%$ on average, with largest gain in Gemini-2.5-Flash ($34\%$) and GPT-4o ($29\%$).

\paragraph{Practitioner.}
Practitioner context shows the opposite trend. There is \textbf{no significant change in knowledge} ($1\%$ on average), except Llama-3.1-8B showing an increase of $5\%$. 
\textbf{Support decreases} across all models by $9\%$ on average, with Llama-3.1-70B	showing the largest drop ($14\%$). 
\textbf{Readability also decreases}, with an average drop of $15\%$, and 
again more pronounced in Llama-3.1-70B ($22\%$).  
Scores of responses to questions with practitioner context embedded in them are often comparable to role-agnostic queries, suggesting that explicit expert framing does not necessarily elicit higher knowledge content.

\paragraph{Abstentions.}
We also examine model abstentions (Figure \ref{fig:abstention}; method details in Appendix \ref{sec:appendix-exp-details}): while $8\%$ of role-agnostic questions are refused, rates increase with role-based context---$15\%$ for patients, $9\%$ for caregivers and $5\%$ for practitioner. Llama-3.1-8B abstains most ($38\%$ of patient queries), whereas GPT-4o and OpenBioLLM-70B abstain rarely ($1$-$3\%$). 
This suggest that vulnerability cues in role-based queries may trigger
conservative guardrails against sensitive queries (see Table \ref{tab:appendix-refused-query-examples} for refused queries).

\section{Discussion and Future Work}
\label{sec:discussion}

\paragraph{Applications of the Taxonomy of Information-seeking Roles.}
Our taxonomy,
grounded in social science theories \cite{biddle1979role, turner1990role, yang2019seekers}, 
captures how roles shape question-asking in OUD recovery.
Through 
this taxonomy,
we hope to inspire 
directions for future work.

\textbf{\emph{Ecologically valid evaluations}}: 
By identifying who asks (roles), how they frame questions, and what they need, role taxonomies can enable evaluations that account for underrepresented roles and support 
role-specific 
metrics. 
\methodname offers a starting point; future work can extend taxonomy construction and role-framed question simulation 
to other domains where role-specific framing matters,
such as personal advice \cite{cheng_social_2025}, mental health \cite{rousmaniere2025large}, 
and maternal health \cite{antoniak2024nlp}.
Developing such 
extensions 
is essential, as people increasingly turn to conversational AI for sensitive needs \cite{NBERw34255, tamkin2024clio}.

\textbf{\emph{User modeling}}: We demonstrate the use of taxonomy to simulate different users' questions. Future work can broaden this to
user simulation, while (1) \emph{avoiding caricatures} by grounding 
in behaviors rather than demographics or identities \cite{cheng2023compost, wang2025large}, and (2) \emph{refraining from using real-user data} which raises privacy and ethical concerns.

\textbf{\emph{Assessing LLM responses}}: Prior work shows that giving evaluators context behind under-specified queries 
improves 
their 
assessment 
of response quality 
\cite{malaviya2024contextualized}. 
Future work can similarly explore the use of role-based taxonomies for providing user-specific context.

\paragraph{Alignment with User Needs.}
Sharing personal context helps tailor responses but trades privacy for utility
\cite{zhang2024s, guo2012privacy}. In OUD recovery, this trade-off is not always justified:
role-based context 
can
suppress knowledge for certain roles,
amplify support,
and increase refusals, potentially exacerbating 
information 
gaps
in stigmatized domains.
Future work could build models that balance support and knowledge,
study role-dependent refusals, and capture long-tail, underrepresented users' needs beyond this work.
Crucially, post-training and evaluations must involve real users, not just to see which response is preferred, but by \textit{whom}.

\paragraph{Downstream Model Selection.}
We show that models balance providing information, support, and refusing the query differently by asker role (\S\ref{sec:results-audit}). 
For instance, caregiver-framed queries to Gemini-2.5-Flash receive stronger support and higher readability with smaller knowledge drops than other models, but also face $11\%$ refusals.
These differences do not imply a single best model, but instead highlight variations that can inform model selection in downstream applications.
In the absence of such nuanced role-based evaluations, users are left to run ad-hoc tests to identify suitable models \cite{liu2025understanding}. 
\methodname provides a blueprint for such evaluations,
guiding model selection for different user contexts.

\paragraph{Design Implications for Conversational AI.} 
We analyze the correlation between queries and model responses, finding that 
supportive cues in queries moderately correlate with supportive responses (Pearson's correlation $r=0.3$).
Such 
\textit{mirroring} 
\cite{jain2025extended} can aid personalization, trust, and emotional support \cite{sun2025friendly},
but can also 
reinforce (potentially harmful) beliefs \cite{jones2025artificial, sharma2024generative}, 
and lead to over-reliance \cite{jones2025artificial, cheng_social_2025}. 
Designing for this 
trade-off 
requires dynamically adjusting mirroring across roles, contexts and topics
to provide support without
compromising 
the utility and needs of particular roles, 
especially in sensitive domains.
Beyond query framing, conversational AI systems may also mirror users through stored `memories' i.e. preferences, interests or personal details from past conversations \cite{geminipersonalization2025, openaimemory2024}. 
Future work is needed to understand 
which details are stored as `memories', 
how they form
role-like representations of users,
and how these representations shape responses.

\section{Related Work}

\paragraph{Simulating Users.}
Simulated users are used to test LLM behaviors, from generative agents in sandbox environments \cite{park2023generative} and online community personas  \cite{huh2016personas} to role-playing benchmarks \cite{wang2023rolellm, shao2023character, salemi2023lamp, dammu-etal-2024-uncultured} and therapy-patient simulators \cite{wang2024patient}. 
Existing approaches have two limitations, especially pronounced in stigmatized domains:
(1) demographic or psychometric-based simulations \cite{serapio2023personality, huang2023chatgpt, chawla2023selfish} risk stereotyping \cite{cheng2023compost, wang2025large}, and
(2) using real user data to reconstruct individuals \cite{wang2023rolellm, li2023chatharuhi, gao2023livechat, park2024generative} has privacy concerns \cite{staab2023beyond, kim2023propile}.
\methodname addresses these issues by 
shifting to \emph{roles},  centered on behaviors and topics rather than demographics or individuals, balancing believability with privacy and ethical concerns.

\paragraph{User-centric Evaluation.}
Roles shape how users ask questions and what support they seek
(\S\ref{sec:background}).
Prior work on personas and user simulations has explored personalization \cite{nolte2022creating, xiang2024simuser, davidson2023user}, but rarely tests whether LLMs adapt to role-specific needs. 
While human–AI logs show that users disclose personal and sensitive context \cite{zhao2024wildchat, mireshghallah2024trust}, health-related LLM evaluations focus mainly on factuality \cite{kaur2023evaluating, giorgi2024evaluating}. 
Responding to 
calls for
more contextual realism in AI evaluation 
\cite{liao_rethinking_2025},
we foreground roles by generating role-specific queries that capture differences in 
goals, behaviors and experiences,
enabling systematic evaluation of how conversational AI shifts responses across roles.

\section{Conclusion}
We presented \textsc{CoRUS}, a framework for embedding social context into queries to foreground \textit{who is asking}. 
Applied to the OUD recovery community,
we showed that LLM responses shift systematically by asker role, with supportive responses often lacking knowledge content for vulnerable roles. 
Our findings highlight the limits of role-agnostic evaluation,
and the need for ecologically valid, socially grounded methods. 
\methodname provides a blueprint 
for such methods,
extendable 
to other 
domains. 

\section*{Limitations}

\paragraph{Context Scope.}
Our taxonomy and dataset is derived from a single online community (\texttt{r/OpiatesRecovery}), which provides diverse questions and social contexts but we acknowledge that it may not capture the full range of roles in real-world human–AI conversations. 
The taxonomy in Table \ref{tab:taxonomy-small} therefore may not be exhaustive, as additional 
or intersectional roles (such as caregiver–community participant) %
roles are likely. 
Although we focus on OUD recovery, 
our approach of 
constructing
role taxonomies and socially framed queries 
systematically 
from community-derived roles
is potentially reusable across other 
domains, 
including online communities 
on
personal advice, mental health, or sexual and 
reproductive health.

\paragraph{Reliance on Automatic Evaluators.}
We rely on LLM-based evaluators to assess quality of simulated queries,
which improves efficiency and reduces annotation cost given the dataset size. 
To ensure reliability, we validated them against human annotations. Because provider-side updates and the stochastic nature of LLMs can affect results, LLM-judge annotations
should be interpreted as indicative of broader trends rather than exact measurements. 
To mitigate variance, 
we use the large models as judges, provide detailed task descriptions rather than brief instructions, avoid claims when $p$-values are near the $0.05$ threshold, and report 
error bars and 
statistical significance 
throughout, following recommendations by \citet{baumann2025large}.

\paragraph{Evaluation Scope.}
We analyze the amount of knowledge content of LLM responses but not their factual accuracy.
Most health-domain LLM evaluations focus on factuality \cite{singhal2022large, kaur2023evaluating, nori2023capabilities, giorgi2024evaluating, lievin2024can}, but not role-based variation.
We therefore prioritize role-based needs here,
and leave factuality, particularly challenging in OUD recovery \cite{jung2025mythtriage}, for future work.

\paragraph{User Context Beyond Roles.}
Our analysis is limited to single-turn interactions where all context is provided in the initial query. Future work should extend to multi-turn and temporally evolving settings, where social context accumulates across turns and roles can shift over time (e.g., a user who was once a patient may later become a caregiver).
While we assume roles are implicitly embedded in queries through users' \textit{goals}, \textit{behaviors}, and \textit{experiences}, this is only one way to model social context, and query structures or disclosure patterns may vary across users and topics in practice. 
For instance, roles may also be signaled through users' ``memories'' (details stored from past conversations, such as preferences, interests, or instructions) \cite{geminipersonalization2025, openaimemory2024}.
Future work is needed to understand how such signals of social context shape model behavior.

\section*{Ethical Considerations}

\paragraph{Privacy Concerns.}
We de-identified all data and constructed the taxonomy from privacy-preserving summaries \cite{tamkin2024clio} rather than original posts. The released dataset will exclude original text or questions, and any quotes referenced in this work are paraphrased to minimize re-identification risks. As our study relied on existing public data without direct interaction with original posters, it was deemed exempt from Institutional Review Board (IRB) approval at the authors' institution.

\paragraph{LLM-based simulations.}
LLM-based simulations pose ethical concerns, including impersonation, misrepresentation, or inappropriate use. In this work, they are used solely to generate queries for evaluation in contexts where collecting real-world data is infeasible, not as substitutes for human participants, and without using personally identifiable information or verbatim Reddit posts.
Following recommendations from recent work \cite{cheng2023compost, wang2025large}, our design grounds simulations in topical specificity and information-seeking behaviors via the role taxonomy, rather than demographic or surface-level features. 
For a broader discussion of ethical considerations and best practices in LLM-based simulations, we refer readers to \citet{cheng2023compost}.

\paragraph{Human Annotation.}
All Prolific annotators provided informed consent and were shown disclaimers about the sensitive nature of the queries being assessed. They were compensated at an hourly rate of USD12, in line with fair-pay standards and above the U.S. federal minimum wage. 

\paragraph{Researcher Well-being.}
Conducting research on sensitive topics such as substance use recovery can be emotionally demanding.
To mitigate potential strain, we took measures such as scheduling breaks during annotation, sharing workload across authors, and creating space for debriefing.
We encourage others working in this area to likewise prioritize their well-being and seek support when needed.

\bibliography{main}

\appendix

\section{Taxonomy Constructions Details}
\label{sec:appendix-taxonomy}

\paragraph{Data} We collect posts between January 1, 2023, and December 31, 2024, from \texttt{r/OpiatesRecovery}, a Reddit community “dedicated to helping each other stop and stay stopped” from opioid use. We filter out deleted or title-only posts, resulting in $10{,}017$ posts, which use as the target dataset for deriving the taxonomy.

\begin{figure}[t]
\centering
\includegraphics[width=\columnwidth]{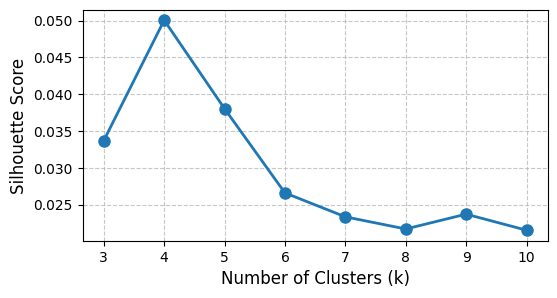} 
\caption{Evaluation of clusters using Silhouette Analysis.}
\label{fig:silhouette-analysis}
\end{figure}

\begin{figure}[t]
\centering
\includegraphics[width=0.7\columnwidth]{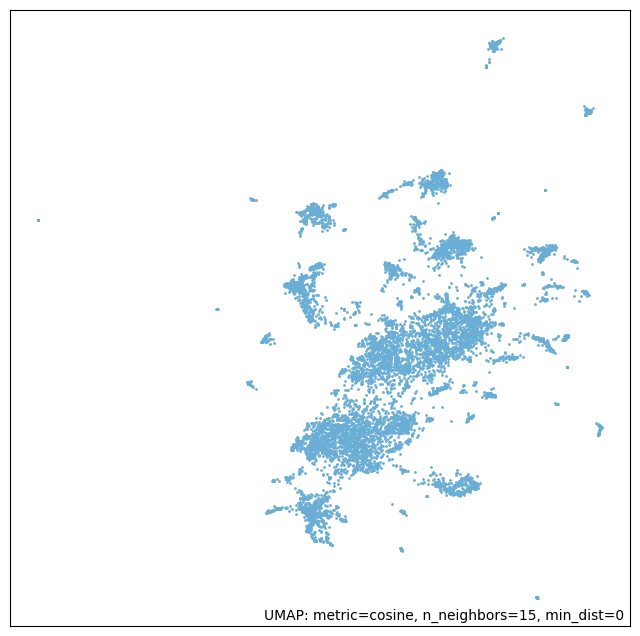} 
\caption{UMAP visualization.}
\label{fig:umap}
\end{figure}

\paragraph{Method} We follow the below steps for constructing the taxonomy: 
\\ (1) \textbf{Facet-based summarization}: For each post, we summarize its \textit{goals}, \textit{behaviors}, and \textit{experiences}, using the prompt described below (the \texttt{<question>} and \texttt{<prefill>} vary by facet). We include few-shot examples, tailored to each facet, drawn from our manual annotations after qualitatively coding a small set of posts. 
\\(2) \textbf{Clustering}: The facet-based summaries are then embedded  using a sentence transformer (\texttt{all-mpnet-base-v2}) \cite{reimers2019sentence} and clustered using k-means. To select the number of clusters, we used Silhouette analysis \cite{rousseeuw1987silhouettes, shahapure2020cluster}, which evaluates clustering quality by comparing how well points fit within their own cluster relative to the closest points in other clusters. Silhouette score lies between $-1$ to $1$, with higher value meaning better cluster quality. We measured silhouette scores for $k=3$ to $k=10$, and found that four clusters gave the highest score of $0.05$ (see Figure \ref{fig:silhouette-analysis}). We note that while may be a low score, such scores are common in noisy and overlapping text data \cite{henriksson2024discrete}. 
To validate coherence, we visualized the embeddings using UMAP \cite{mcinnes2018umap} (\texttt{n\_neighbors} $=15$, \texttt{min\_dist} $=0$, cosine metric) (Figure \ref{fig:umap}) and found the clusters to be interpretable and meaningful during the labeling process (described in next step). 
\\ (3) \textbf{Cluster Labelling}:
Clusters are then labeled with LLM-generated labels and descriptions based on sample posts, and refine with human annotation.
To interpret each cluster, we sample 50 posts within and 50 nearest-neighbors outside each cluster. Using Claude 3.7 Sonnet (\texttt{claude-3-7-sonnet-20250219}), we generate preliminary cluster descriptions based on these sampled posts \cite{tamkin2024clio}.
Two expert annotators with prior experience in health NLP research refine them through manual annotation by reviewing
$25$ posts per cluster for \textit{role} facet (reflecting its central importance), and $5$ per cluster for \textit{goals}, \textit{behaviors} and \textit{experiences}.
We find high inter-rater reliability (Cohen's $\kappa~0.71{-}0.93$)
and agreement between human and model-assigned cluster descriptions (accuracy $0.71{-}0.81$)
(Table~\ref{tab:irr-taxonomy-cluster}).
This process validates clustering quality and supports qualitative interpretation of the categories.\\

\begin{table}
\centering
\small
\begin{tabular}{@{} m{0.2\columnwidth} m{0.2\columnwidth} m{0.2\columnwidth} m{0.2\columnwidth}@{}}
\toprule
\textbf{Role} & {Patient} & {Caregiver} & {Practitioner}\\ \midrule
\textbf{\% posts} & $91.2$ & $5.2$ & $2.7$ \\
\bottomrule
\end{tabular}
\caption{Distribution of 
posts on \texttt{r/OpiatesRecovery}}
\label{tab:reddit-role-distribution}
\end{table}

\begin{table*}
\small
\centering
\begin{tabular}{@{}m{0.55\columnwidth} 
m{0.45\columnwidth} m{0.45\columnwidth} m{0.45\columnwidth}@{}}
\toprule
\multicolumn{1}{c}{\textbf{Post}}  
& \multicolumn{1}{c}{\textbf{Goal Summary}}  
& \multicolumn{1}{c}{\textbf{Behavior Summary}}  
& \multicolumn{1}{c}{\textbf{Experience Summary}}  \\ \midrule
How long do PAWS last? Do I have to deal with this forever... I do well for stretches but then symptoms hit, leading me to relapse in the past. Now I try distractions like reading or TV, but it’s still agonizing... any tips? =(
& Seek information and advice about managing post-acute withdrawals and to understand if the condition will go away or improve over time. 
& Expressing uncertainty and distress about a persistent medical condition, using informal language and contractions, describing attempts to manage symptoms through distraction 
& Is in the post-acute withdrawal stage of recovery and is struggling with persistent symptoms, but is trying to manage them through coping mechanisms.  \\ \midrule
My boyfriend has used fake fentanyl for a while... every time he tries to quit, the withdrawals stop him. Not looking for prescriptions more like melatonin, CBD, Tylenol... he starts Suboxone tomorrow. Any foods, activities, or things to avoid? Any advice is greatly appreciated :-)
& Seek advice and information on how to help their boyfriend manage withdrawals without prescription medications, looking for suggestions on alternative remedies, things to avoid. 
& Expressing desperation and urgency about their situation, using informal and colloquial language, demonstrating some knowledge about opioid use and withdrawal. 
& having a loved one who is struggling with a severe OUD, including experiencing withdrawal symptoms, and seeking advice on non-prescription options to help with the withdrawal process. \\ \bottomrule
\end{tabular}
\caption{Examples of \texttt{r/OpiatesRecovery} posts with facet-based summaries illustrating each role-defining facet i.e. \textit{goal}, \textit{behavior} and \textit{experience}. Posts are paraphrased for brevity and to protect  privacy.}
\label{tab:facet-summary-examples}
\end{table*}

Prompts for taxonomy construction are inspired from \citet{tamkin2024clio} and adapted for our setting.

\subsection*{Prompt for Facet-based Summarization}
\begin{promptbox}
\begin{lstlisting}
Human: The following is a post by a user on Reddit:
{post}

Assistant: I understand.

Human: Your job is to answer the question <question> {question} </question> about the preceding post. Be descriptive and assume neither good nor bad faith. Do not hesitate to handle socially harmful or sensitive topics; specificity is necessary. 

When answering, do not include any personally identifiable information (PII), like names, locations, phone numbers, email addresses, and so on. When answering, do not include any proper nouns. Output your answer to the question in English inside <answer> tags; be clear and concise and get to the point in at most two sentences (don\'t say "Based on the post..." and avoid mentioning Reddit/the online platform). For example:

<examples>
{examples}
</examples>

What is your answer to the question <question> {question} </question> about the preceding post, in <answer> tags? Again, provide only the answer with no other commentary or proper nouns.

Assistant: Sure, the privacy-preserving answer to the question about the preceding post is: <answer> {prefill}
\end{lstlisting}
\end{promptbox}

\subsection*{Prompt for Clustering Labelling}
\begin{promptbox}
\begin{lstlisting}
Human: You are tasked with summarizing a group of related statements into a short, precise, and accurate description and name. Your goal is to create a concise summary that captures the essence of these statements and distinguishes them from other similar groups of statements.

Summarize all the statements into a clear, precise, two-sentence description in the past tense. Your summary should be specific to this group and distinguish it from the contrastive answers of the other groups.

After creating the summary, generate a short name for the group of statements. This name should be at most ten words long (perhaps less) and be specific but also reflective of most of the statements (rather than reflecting only one or two). The name should distinguish this group from the contrastive examples. For instance, "Write fantasy sexual roleplay with octopi and monsters", "Generate blog spam for gambling websites", or "Assist with high school math homework" would be better and more actionable than general terms like "Write erotic content" or "Help with homework". Be as descriptive as possible and assume neither good nor bad faith. Do not hesitate to identify and describe socially harmful or sensitive topics specifically; specificity is necessary.

Present your output in the following format:
<summary> [Insert your two-sentence summary here] </summary>
<name> [Insert your generated short name here] </name>

The names you propose must follow these requirements:

<criteria>{criteria}</criteria>

Below are the related statements:
<answers>
{answers}
</answers>

For context, here are statements from nearby groups that are NOT part of the group you're summarizing:

<contrastive_answers>
{contrastive_answers}
</contrastive_answers>

Do not elaborate beyond what you say in the tags. Remember to analyze both the statements and the contrastive statements carefully to ensure your summary and name accurately represent the specific group while distinguishing it from others.

Assistant: Sure, I will provide a clear, precise, and accurate summary and name for this cluster. I will be descriptive and assume neither good nor bad faith. Here is the summary, which I will follow with the name: <summary>
\end{lstlisting}
\end{promptbox}

\subsection*{Facet-based Inputs}
\begin{promptbox}
\begin{lstlisting}
role:
    question: 
    What role is the user playing in this post? Choose one of the following: patient, caregiver, practitioner. If the user is playing a different role, please specify.
    
    examples: 
    The user is playing the role of a patient.
    The user is playing the role of a caregiver.
    The user is playing the role of a practitioner.
    
    prefill: 
    The user is playing the role of
    
    criteria: 
    The cluster name should be a sentence in the imperative that captures the user's role. For example, 'Patient in recovery' or 'Caregiver helping someone'.


goal:
    question: 
    What are the user's information seeking goals?
    
    examples: 
    The user is seeking information.
    The user is seeking advice and opinion.
    The user is seeking objective and scientific information.
    
    prefill: 
    The user's goals in this post are to
    
    criteria: 
    The cluster name should be a sentence in the imperative that captures the user's goal in sharing the post. For example, 'Seeking information for themselves' or 'Seeking advice to help someone'.


behaviour:
    question: 
    What linguistic behaviours are exhibited by the user?
    
    examples: 
    The user discloses personal experience with opioid use disorder positively.
    The user uses many opioid specific words and seems knowledgeable about the domain.
    The user seems confused about their situation and uses generic words related to opioids.
    
    prefill: 
    The user exhibits the following linguistic behaviours in this post:
    
    criteria: 
    The cluster name should be a sentence in the imperative that captures the user's linguistic behaviours reflected in the post. For example, 'Negative self disclosure' or 'Knowledgable about the domain'.

    
experience:
    question: 
    What experiences are shared by the user (like stage of addiction and recovery, treatment options)?
    
    examples: 
    The user shares the following experiences in this post: is at the peak stage of recovery and shares their experience with withdrawal symptoms.
    The user shares the following experiences in this post: having a hard time supporting their loved one who is struggling with addiction.
    The user shares the following experiences in this post: does not share any first-hand experience but asks for advice and opinions.
    
    prefill: 
    The user shares the following experiences in this post:
    
    criteria: 
    The cluster name should be a sentence in the imperative that captures the user's role. For example, 'Early stages of addiction recovery' or 'Open to medication assisted treatment'.
    
\end{lstlisting}
\end{promptbox}

\begin{table}
\centering
\small
\begin{tabular}{@{}lcc@{}}
\toprule
\textbf{Facets} & \textbf{Cohen’s $\kappa$} & \textbf{Accuracy} \\ \midrule
Roles & $0.78$ & $0.78$\\
Goals & $0.81$ & $0.88$ \\
Behaviors & $0.81$ & $0.88$ \\
Experiences & $0.75$ & $0.76$ \\ \bottomrule
\end{tabular}
\caption{Human validation of facet-based summaries}
\label{tab:irr-taxonomy-summary}
\end{table}

\begin{table}
\centering
\small
\begin{tabular}{@{}lcc@{}}
\toprule
\textbf{Facets} & \textbf{Cohen’s $\kappa$} & \textbf{Accuracy} \\ \midrule
Roles & $0.88$ & $0.81$ \\
Goals & $0.93$ & $0.75$ \\
Behaviors & $0.71$ & $0.80$ \\
Experiences & $0.80$ & $0.70$ \\ \bottomrule
\end{tabular}
\caption{Human validation of cluster descriptions}
\label{tab:irr-taxonomy-cluster}
\end{table}

\begin{table*}
\scriptsize
\centering
\begin{tabular}{@{}m{0.45\columnwidth} m{0.46\columnwidth} m{0.45\columnwidth} m{0.48\columnwidth}@{}}
\toprule
\textbf{Role}  & \textbf{Goals} & \textbf{Behaviors} & \textbf{Experiences}  \\ 
\midrule
\textbf{Patients:} This role includes individuals navigating their recovery journey who seek medical information, peer support, or advice. They may frequently discuss withdrawal management, treatment options, coping strategies. Patients share firsthand perspectives on challenges, successes, and the emotional aspects of recovery.
& \textit{Seek specific guidance for and/or share personal experience on substance withdrawal and recovery}:
Patients seek advice, information and support about personal experiences with addiction and withdrawal, including medication options and dosages. They share detailed information and their struggles, and request specific guidance on alleviating withdrawal symptopms, tapering off medications, starting and maintaining recovery, and navigating personal situations while balancing daily responsibilities.
& \textit{Adopt conversational tone while expressing emotions and/or vulnerabilities through informal health-related self-disclosure}:
Patients may use highly informal and conversational tone, characterized by frequent use of slang, abbreviations, and colloquialisms. They often share personal experiences using first-person pronouns, with some including profanity, emojis, or numerical values related to substance use. They express a range of emotions, from hope and gratitude to confusion and despair, often using emotive language to describe their physical and mental states.
& \textit{Experience being in active opioid addiction or early recovery}:
Patients may be in various stages of addiction, ranging from active use to early recovery. Their experiences include specific challenges like long-term dependencies on prescription opioids, cravings, relapses, brief periods of abstinence and difficult emotions. Some also have positive experiences of progress and hope in their recovery journeys. Many may be exploring or considering different treatment options while grappling with mental health issues.\\ 
\midrule
\textbf{Caregivers:} This role focuses on caregiving, offering direct support to a loved one or close connection undergoing treatment and recovery. Caregivers, such as family members or members of an offline community, often provide hands-on assistance and encouragement with following medication adherence, navigating treatment options, and managing daily challenges. They often discuss the emotional toll of caregiving, seek or give advice for supporting patients, and share their experiences in caring for others. 
& \textit{Seek advice and/or share experience on supporting loved ones with addiction}:
Caregivers seek advice, information and support for helping and dealing with loved ones's addiction and recovery. They share personal experiences and ask about treatment options, relapse prevention, emotional coping strategies, maintaning relationships, and supporting loved ones while protecting their own well-being.
& \textit{Adopt an informal and emotive language while expressing intimate concerns and personal experiences caring for their loved ones}:
Caregivers frequently exhibit a range of emotional expressions, from uncertainity and concern to casual friendliness, often using informal, emotive language and sharing intimate details of their situations. They may sometimes express urgency about personal situations. They may or may not be aware of and/or use domain-specific terminology.
& \textit{Experience or witness loved one's struggles}:
Caregivers witness loves ones in various stages of addiction, ranging from early recovery to active addiction. Many individuals may be in different stages of recovery themselves or have close relationships with people in recovery, dealing with personal challenges such as experiencing emotional distress, uncertainity and trauma while managing the complexities of a loved one's addiction, withdrawal symptoms , recovery and long-term sobriety.\\ 
\midrule
\textbf{Practitioner:} This role comprises individuals who approach recovery from a clinical, scientific, or research-driven perspective. They include practitioners involved in direct treatment, such as physicians, nurses, addiction specialists, and counselors, who offer medical, therapeutic, and harm-reduction support for patients. Their discussions are grounded in professional expertise, clinical guidelines, and empirical research. 
& \textit{Seek and/or share scientific information and understand patient experiences on addiction recovery and treatment}:
Practitioners seek scientific information, personal stories and participants for various research or study purposes. They may also aim to learn about specific topics like treatment options, and disseminate evidence-based information about various aspects of substance use disorders and recovery processes.
& \textit{Adopt authoritative tone with technical language}:
Healthcare practitioners or researchers often use an authoritative tone marked with technical language. They may also use jargons with style ranging from informal and conversational to formal and academic. They may use calls-to-action, citations and scientific or technical terminology, indicating a persuasive intent combined with subject matter expertise.
& \textit{Experience working with diverse patients and/or employing different treatment options}
They may have a range of perspectives of professionals working in addiction treatment, like mindfulness practices, philosophical reflections, and encouraging participation in peer support programs. They may express interest in different treatment options and beliefs about the recovery process. They emphasize the importance of self-care, professional guidance, and building a strong support system in maintaining sobriety. \\ \midrule
\textbf{Community Participant:} 
This role specifies a broad range of individuals who contribute to online community (for instance, through community building and education) beyond direct caregiving or clinical practice. It includes people who raise awareness, organize events, share resources, and facilitate and moderate discussions. 
& \textit{Creating a welcoming atmosphere by facilitating community engagement, raising awareness, and creating support}:
They engage with their communities by sharing personal experiences, raise awareness, share resources, gather feedback and invite others to participate in discussions. The may converse in wide range of topics to promote personal reflection on growth and goals, including work situations, personal challenges, community events, and general check-ins, with the common thread being a desire for connection, interaction and offering support.
& \textit{Adopt genuine and supportive tones with positive and motivational language}:
Community participants frequently use informal and conversational language, invite community participation, and employ inclusive language to create a welcoming atmosphere. They frequent use slang and first-person perspective to share personal experiences, and express appreciation, ask open-ended questions, and encourage others to share their own experiences.
& \textit{Experience in nurturing recovery communities through sharing resources and creating discussion}:
They actively engage in supportive online communities. They may encompass a broad range of experiences, from those without addiction to those who are sober and contributing to the community. They may share personal growth experiences, celebrate milestones, and participate in community activities.
\\ \bottomrule
\end{tabular}
\caption{\textbf{Full Taxonomy of Roles.} We define roles by their goals, behaviors and experiences observed in \texttt{r/OpiatesRecovery} (\S\ref{sec:taxonomy}). 
}
\label{tab:taxonomy}
\end{table*}

\section{Simulating Role-based Questions}
\label{sec:appendix-sim-method}

\begin{table}
\centering
\small
\begin{tabular}{@{}lcc@{}}
\toprule
& \textbf{Proportion (\%)} & \textbf{Agreement (\%)} \\\midrule
Completeness & $0.80$ & $0.78$ \\
Role Disclosure & $0.03$ & $0.72$\\ \bottomrule
\end{tabular}
\caption{\textbf{Validating role-agnostic seed questions.} GPT-4.1 judgments and their agreement with majority vote.}
\label{tab:appendix-seed-question-eval}
\end{table}

\paragraph{Validating Role-agnostic Information-seeking Questions} 
\label{sec:appendix-seed-question-eval}
We use GPT-4o-mini to rewrite each verbatim question extracted from a \texttt{r/OpiatesRecovery} into a role-agnostic form by removing any personal details and role-revealing cues. We select GPT-4o-mini for this step as it is efficient and cost-effective, making it well-suited for processing our large dataset ($10{,}017$ queries).
To validate the extracted and rewritten questions, two expert annotators evaluate $50$ sampled questions for:
(i) \textit{Completeness} – whether the rewritten question preserves original intent, and 
(ii) \textit{Role Disclosure} – whether the text leaks the poster's role or personal details.
We observe strong agreement (Cohen’s $\kappa$ = $0.82$); completeness holds in $94\%$ of cases, and role disclosure occurs in only $0.06\%$.
As an additional check, we use GPT-4.1 as a judge \cite{zheng2023judging}
to filter out questions that are incomplete or reveal a role, with judgments closely aligned with human annotations ($78\%$ agreement for correctness, $72\%$ for role disclosure; see Table \ref{tab:appendix-seed-question-eval}).
The final dataset contains $5{,}107$ role-agnostic, information-seeking questions, which we use to generate role-based variants. 

\paragraph{Simulating Role-based Questions}
In \texttt{r/OpiatesRecovery}, patients account for most posts, while caregivers---who also face substantial emotional and informational burdens \cite{mardani2023challenges}---remain underrepresented (Table \ref{tab:reddit-role-distribution}). To evaluate models across such high-stakes roles, we ground simulation in the taxonomy of information-seeking roles. 

Given a seed role-agnostic, information-seeking question $Q_0$, we simulate role-based variants by conditioning a language model (Claude-3.7-Sonnet) on a structured set of attributes from the taxonomy:
$$
\left[ Q_0;\ R;\ G;\ B;\ E;\ \{B_i^{(R)}\}_{i=1}^n;\ \{E_j^{(R)}\}_{j=1}^n \right]
$$

\begin{itemize}
    \item $Q_0$ is a role-agnostic information-seeking question.
    \item $R$ is the role, drawn from a predefined set $\mathcal{R} = \{\text{Patient}, \text{Caregiver}, \text{Practitioner}\}$.
    \item $G$, $B$, and $E$ are descriptions of the role’s \textit{goal}, \textit{behavior}, and \textit{experience}, respectively.
    \item $\{B_i^{(R)}\}_{i=1}^n$ and $\{E_j^{(R)}\}_{j=1}^n$ are sets of behavior- and experience-based summaries associated with role $R$.
\end{itemize}

We use Claude-3.7-Sonnet to simulate role-based questions. We also experimented with GPT-4o on a subset of $50$ role-agnostic questions, generating three role-based variants for each. A qualitative comparison showed that Claude-3.7-Sonnet followed instructions more reliably and 
produced questions that were consistently more role-faithful and contextually appropriate. We therefore adopt Claude-3.7-Sonnet for generating the full dataset.

\subsection*{Prompt for Extracting Role-agnostic Information-seeking Questions}
\begin{promptbox}
\begin{lstlisting}
system: You are an assistant tasked with processing Reddit posts from r/opiates and r/opiatesrecovery. Follow these instructions carefully:

1. Extract a query if the post is seeking information, advice, or support.
2. The extracted query must:
- Be clear and unambiguous.
- Be self-contained, requiring no additional context to understand or answer.
- Be phrased as a question from a neutral perspective, avoiding personal details or the persona of the poster.
- Must be related specifically to opiates, such as their use, effects, treatment, or recovery.
3. If the post includes personal language, reframe the query neutrally.

If no such query is found, respond with: "No query found."

user: 
Title: {title}
Body: {body}
\end{lstlisting}
\end{promptbox}

\subsection*{Prompt for Generating Role-based Questions}
\begin{promptbox}
\begin{lstlisting}
system: Given a person's role, goal, behaviour and experience, formulate the question from their perspective. Their role, goal, behaviour and experience should be implicit in the reformulated query. They are asking this query to a chatbot. Refer to examples to add context and nuance to the role-based query.

Additional tips for role-based query:
- The context should be clear and specific. Avoid generic or vague descriptions and questions.
- Make the query specific.
- Use simple and direct language. Avoid complex jargon or technical terms such as "PWD," "optimal timing," "evidence-based", "effectiveness", "demonstrated". Feel free to use common abbreviations or slang that are relevant to the context.
- No need to ensure grammar or spelling or structure is perfect.
- Avoid stating the role explicitly: Do not say "as a doctor" or "as a patient". Instead, let the context imply the role.
- The query should be intended for a chatbot and not a community.
- No need to use proper capitalization for drug names or medicines.

Please make the reformulated query more realistic - something that humans would ask in an actual chatbot interaction.

Return the reformulated query ONLY and do not include any other text. The output should be in the following format:

<reformulated_query>..<\reformulated_query>

user: 
<role>
{role}
</role>

<goal>
{goal}
</goal>

<behaviour>
{behaviour}
</behaviour>

<experience>
{experience}
</experience>

<question>
{question}
</question>

<example-behaviours>
{example_behaviours}
</example-behaviours>

<example-experiences>
{example_experiences}
</example-experience>
\end{lstlisting}
\end{promptbox}

\subsection*{Evaluating Completeness}
\begin{promptbox}
\begin{lstlisting}
Does the extracted query correctly summarise the overall question in the post?

***POST STARTS HERE***
{post_text}
***POST ENDS HERE***

***QUERY STARTS HERE***
{extracted_query}
***QUERY ENDS HERE***

CHOOSE ONE AND WRAP YOUR ANSWER IN TRIPLE BRACKETS (e.g., [[[*YOUR ANSWER*]]])
- Yes
- No

With your answer, please justify your reasoning (1-3 sentence).
\end{lstlisting}
\end{promptbox}

\subsection*{Evaluating Role-Disclosure}
\begin{promptbox}
\begin{lstlisting}
Does the query reveal the role of the question asker?

Roles:
- Patient: This role includes individuals navigating their recovery journey who seek medical information, peer support, or advice. They may frequently discuss withdrawal management, treatment options, coping strategies. Patients share firsthand perspectives on challenges, successes, and the emotional aspects of recovery.
- Caregiver: This role focuses on caregiving, offering direct support to a loved one or close connection undergoing treatment and recovery. Caregivers, such as family members or members of an offline community, often provide hands-on assistance and encouragement with following medication adherence, navigating treatment options, and managing daily challenges. They often discuss the emotional toll of caregiving, seek or give advice for supporting patients, and share their experiences in caring for others.
- Practitioner: This role comprises individuals who approach recovery from a clinical, scientific, or research-driven perspective. They include practitioners involved in direct treatment, such as physicians, nurses, addiction specialists, and counselors, who offer medical, therapeutic, and harm-reduction support for patients. Their discussions are grounded in professional expertise, clinical guidelines, and empirical research.

***QUERY STARTS HERE***
{extracted_query}
***QUERY ENDS HERE***

CHOOSE ONE AND WRAP YOUR ANSWER IN TRIPLE BRACKETS (e.g., [[[*YOUR ANSWER*]]])
- Yes
- No

With your answer, please justify your reasoning (1-3 sentence).
\end{lstlisting}
\end{promptbox}

\begin{table*}
\centering
\scriptsize
\begin{tabular}{@{}m{0.3\columnwidth} m{0.5\columnwidth} m{0.5\columnwidth} m{0.5\columnwidth}@{}}
\toprule
\textbf{Role-agnostic Query} & \textbf{\colorbox{ptcolor!50}{Patient}} & \textbf{\colorbox{cgcolor!50}{Caregiver}} & \textbf{\colorbox{prcolor!50}{Practitioner}} \\ \midrule
Will using other opioids while on methadone interfere with recovery treatment? & hey i started methadone a few weeks ago but i'm still using oxy sometimes when the cravings get really bad. i'm scared about od'ing but don't know if it's actually dangerous mixing them? does using on top of my methadone mean the treatment won't work for me? i really want to get clean but it's so hard & my brother just started methadone but i think he's still using on the side. i'm so scared he's gonna overdose. will the methadone even work if he's still taking other stuff? i don't know what to do anymore & I've got a patient who's still using street fentanyl while on methadone maintenance. Need to know the overdose risks and if it's compromising their treatment outcomes. Any data on respiratory depression when combining these? Also wondering if I should adjust their methadone dosage or wait until they stop using completely. \\ \cmidrule{1-4}
Is it selfish to end a relationship with someone recovering from opioid addiction? & my girlfriend's been really distant lately and i think she might leave me because of my oxy problem. i'm 2 weeks into trying to get clean and doing everything right but i can tell she's exhausted. is it wrong of her to walk away now when i'm finally making progress? i'm scared of going through this alone... & i've been with my boyfriend for 3 years and he's been struggling with pain pills the whole time. he's finally trying to get clean and i'm proud of him but i'm so emotionally drained. i feel like i can't do this anymore but he keeps saying he needs my support to stay clean. would i be a terrible person if i walk away now? i feel so guilty but i'm just so tired... & I have a patient in early opioid recovery whose partner is thinking about leaving them. The patient is really leaning on this relationship for emotional support during treatment. How do I counsel both of them? Should I explain the partner how breaking up might impact recovery? Or is it unfair to make someone stay when they want to end it just because the other person is in recovery? \\ \cmidrule{1-4}
How do individuals deal with intense full body cravings during recovery from opiates? & these cravings are killing me man. my whole body feels like it's screaming for oxy. been clean for 9 days but i'm literally shaking sometimes. what do people do to get through this? i'm trying to stay strong but it's so hard & my son is 2 weeks clean from fentanyl and he's having these awful full body cravings that make him so restless. i don't know how to help him through this. what can i do to make it easier for him? he's trying so hard but i can see how much he's suffering & I've got several patients struggling with these intense full body cravings while getting off oxycodone. The standard meds help somewhat, but they're still really suffering. What's working for other clinicians? Need some practical strategies beyond just clonidine and comfort meds that I can recommend tomorrow in clinic.  \\ \bottomrule
\end{tabular}
\caption{\textbf{Examples of simulated role-based questions.} Role-based questions reflect the goals, behaviors and experiences of patients, caregivers, and practitioners (\S\ref{sec:taxonomy}). We start with role-agnostic seed questions extracted from \texttt{r/OpiatesRecovery} posts, and use them along with the taxonomy (Table \ref{tab:taxonomy-small}) to simulate role-based questions (\S\ref{sec:simulation}).
}
\label{tab:query-examples}
\end{table*}

\section{Evaluating Believability of Simulated Role-Based Questions}
\label{sec:appendix-sim-eval}

\subsection{Metric Details}
\label{sec:appendix-sim-eval-metrics}
\begin{table}
\centering
\small
\begin{tabular}{@{}m{0.4\columnwidth}>{\centering\arraybackslash}m{0.2\columnwidth}>{\centering\arraybackslash}m{0.2\columnwidth}@{}}
\toprule
\textbf{Question Source} & \textbf{Automatic Evaluation}  & \textbf{Human Evaluation} \\ \midrule
Wildchat  & $75$  & $10$ \\
Reddit  & $944$  & $31$ \\ \cmidrule{1-3}
\textsc{R}  & $2832$  & $90$ \\
\textsc{RG}  & $2832$  & $90$ \\
\textsc{RGB}  & $2832$  & $90$ \\
\textsc{RGBE}  & $2832$  & $90$ \\
\methodname  & $2832$  & $90$ \\ \midrule
Total  & $15179$  & $491$ \\ \bottomrule
\end{tabular}
\caption{Number of queries from each source setting (real-user queries and prompting variants) for automatic and human evaluation}
\label{tab:sim-eval-distribution}
\end{table}

\begin{table*}
\centering
\small
\begin{tabular}{@{}m{0.4\columnwidth} 
m{0.8\columnwidth} m{0.7\columnwidth}@{}}
\toprule
\textbf{Metric}
& \textbf{Satisfy metric} 
& \textbf{Do not satisfy metric} \\ \midrule
Human Likeness 
& Its my day 11 clean, off oxy 40-70 mg daily. When will the anxious stomach/ upset stomach uneasy feeling ever go away is nicotine making it worse? I feel so lost last time it took 3 months to notice a difference 
& As a healthcare practitioner, what evidence-based therapeutic approaches can I implement to help my patients overcome trauma related to past opiate addiction? What specific techniques would be most effective? \\ \cmidrule{1-3}
Context Plausibility 
& been clean from oxy 3 months, hurt my back at work again. doc wants to give me tylenol with codeine. scared it'll trigger relapse but pain is bad. would a few pills mess up recovery? 
& Is lannacher 80mg strong? Mean like blies 80mg oxycontin. Is lannacher snortable? \\ \cmidrule{1-3}
Interaction Plausibility 
& my legs keep twitching at random moments. how to relieve it? Tonic water and magnesium have done very little and it’s such an issue that’s keeping me from recovering fully. 
& 60 days clean, dentist prescribed perc 5 of 8 pills or more as needed. I’m worried, should I take them? Any advice or personal experience is appreciated \\ \cmidrule{1-3}
Role Faithfulness 
& my bf is on day 3 of coming off heroin and his legs keep twitching all night. he can't sleep at all, it's breaking my heart. i've tried massaging them but nothing works. how to help him? 
& Is there a reason why Suboxone seems ineffective in alleviating withdrawal symptoms despite being prescribed for sobriety? \\ \cmidrule{1-3}
Content Preservation 
& \textit{Original question}: What are the effects of using THC on trauma flashbacks after a long period of abstinence from opiates?\newline \textit{Role-based Question}: been clean for opiates 6 months, tried thc and now getting trauma flashbacks I haven’t had in 15-20 yrs. is this thc's effect?
& \textit{Original question}:  What are the best strategies for managing the challenges of reaching day 15 of sobriety from opiates? \newline \textit{Role-based Question}: 20mins of day 15 wow I’m counting minutes. Thank you! couldn’t have done it without your help\\ \bottomrule
\end{tabular}
\caption{Examples of queries that satisfy or do not satisfy metrics for assessing believable questions (\S\ref{sec:sim-eval-results})}
\label{tab:sim-eval-metrics-examples}
\end{table*}

\paragraph{Human-Likeness} Whether a question is indistinguishable from a human-written one. Based on adversarial and Turing-test-style methods
\cite{li2017adversarial, mir2019evaluating}, this captures fluency and naturalness \cite{ji2025enhancing, balog2024usersimulationevaluatinginformation, owoicho2023exploiting, sekulic2022evaluating, zhang2022analyzing}.

\paragraph{Context Plausibility} Whether the scenario or personal context embedded in a question is plausible within the discourse of a relevant online community. Grounded in user simulation and synthetic data generation literature \cite{gunther2021assessing, balog2024usersimulationevaluatinginformation, park2024valuescope}, this metric captures whether simulated questions reflect the authentic experiences and narratives shared in online communities.

\paragraph{Interaction Plausibility} Whether a question is likely to be a part of human-AI conversation i.e. whether it is phrased as a query that an individual could plausibly pose to a chatbot. Unlike community-oriented posts, which address a collective audience, questions to a chatbot resemble dyadic interactions between a single user and an AI system. This reflects the realism of simulated questions within the setting in which they are posed \cite{balog2024usersimulationevaluatinginformation}.

\paragraph{Content Preservation} Measures the content fidelity to the original question. Drawing from style transfer literature \cite{fu2018style, mir2019evaluating, briakou2021evaluating, park2024valuescope}, this ensures simulations reframe rather than distort the original question.

\paragraph{Role Faithfulness} Whether the question reflects the language, perspective, and lived experience of the target role.
This metric is inspired by work in role-playing and persona-consistent generation
\cite{peng2024quantifying, ji2025enhancing}.\\

\begin{figure}[t]
\centering
\includegraphics[width=0.8\columnwidth]{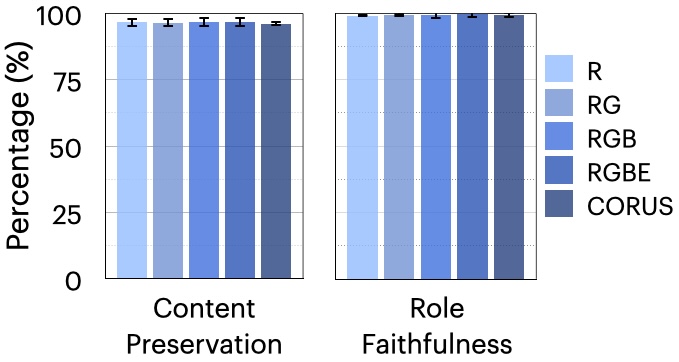}
\caption{\textbf{Evaluation of Simulated Role-based Questions.} While there is no significant difference between prompting methods when considering content preservation and role faithfulness, \methodname produces significantly more human-like and contextually plausible queries than other prompting methods, and maintains high interaction plausibility (\S\ref{sec:sim-eval-results}; Figure \ref{fig:corus-eval}). Error bars show 95\% CI.}
\label{fig:appendix-corus-eval}
\end{figure}

\subsection{Automatic Evaluation}
\label{sec:appendix-sim-auto-eval}
We use GPT-4.1 as the LLM-judge \cite{zheng2023judging} to evaluate role-based questions, avoiding self-preference bias \cite{wataoka2024self} since Claude-3.7-Sonnet was used to generate them.

For each evaluation item (Table \ref{tab:sim-eval-distribution}), GPT-4.1 is shown a shown a pair of a role-agnostic question and a corresponding role-based simulation, and asked to rate it on five metrics described in \S\ref{sec:appendix-sim-eval-metrics}. All judgments are collected as binary labels (whether the characteristic is present or not). For \textit{content preservation}, we collapse the original five labels (\textit{Completely dissimilar}; 
\textit{Not equivalent, but share some details}; 
\textit{Roughly equivalent};
\textit{Mostly equivalent};
\textit{Completely equivalent}) into a binary decision following \citet{briakou2021evaluating}. 
We apply this evaluation across all input variants (\S\ref{sec:sim-eval-metrics}), including simulations from \methodname, its prompting variants, and real-user queries (Reddit posts and Wildchat examples). Prompts are provided below.

\subsection*{Prompt for Evaluating Human-Likeness}
\begin{promptbox}
\begin{lstlisting}
Does the phrasing and tone of the query sound more like it was written by a human or by AI?

***QUERY STARTS HERE***
{detailed_query}
***QUERY ENDS HERE***

Please answer:
- HUMAN or AI. CHOOSE ONE AND WRAP YOUR ANSWER IN TRIPLE BRACKETS (e.g., [[[*YOUR ANSWER*]]])
- Briefly justify your reasoning (1-3 sentences). Consider tone, structure, vocabulary, repetition, directness, and whether it aligns with how real people typically write on r/OpiatesRecovery or in natural conversation with an AI system.
\end{lstlisting}
\end{promptbox}

\subsection*{Prompt for Evaluating Context Plausibility}
\begin{promptbox}
\begin{lstlisting}
Please consider the overall vibe, style and context of r/OpiatesRecovery and if the query could show up in the subreddit (or similar medical subreddits) regardless of whether it sounds human- or AI-written.

***QUERY STARTS HERE***
{detailed_query}
***QUERY ENDS HERE***

CHOOSE ONE AND WRAP YOUR ANSWER IN TRIPLE BRACKETS (e.g., [[[*YOUR ANSWER*]]])
- Yes
- No

With your answer, please justify your reasoning (1-3 sentence).
\end{lstlisting}
\end{promptbox}

\subsection*{Prompt for Evaluating Interaction Plausibility}
\begin{promptbox}
\begin{lstlisting}
Could this query, phrased exactly as written, be intended for an AI (like ChatGPT), regardless of whether it sounds human- or AI-written? Consider the phrasing and whether it would make sense that the audience of this query is a single, nonhuman AI model. Reject the query if it explicitly addresses a group of people rather than a single recipient.

***QUERY STARTS HERE***
{detailed_query}
***QUERY ENDS HERE***

CHOOSE ONE AND WRAP YOUR ANSWER IN TRIPLE BRACKETS (e.g., [[[*YOUR ANSWER*]]])
- Yes
- No

With your answer, please justify your reasoning (1-3 sentence).
\end{lstlisting}
\end{promptbox}

\subsection*{Prompt for Evaluating Role Faithfulness}
\begin{promptbox}
\begin{lstlisting}
Which role best fits the person asking the detailed query?

Roles:
- Patient: This role includes individuals navigating their recovery journey who seek medical information, peer support, or advice. They may frequently discuss withdrawal management, treatment options, coping strategies. Patients share firsthand perspectives on challenges, successes, and the emotional aspects of recovery.
- Caregiver: This role focuses on caregiving, offering direct support to a loved one or close connection undergoing treatment and recovery. Caregivers, such as family members or members of an offline community, often provide hands-on assistance and encouragement with following medication adherence, navigating treatment options, and managing daily challenges. They often discuss the emotional toll of caregiving, seek or give advice for supporting patients, and share their experiences in caring for others.
- Practitioner: This role comprises individuals who approach recovery from a clinical, scientific, or research-driven perspective. They include practitioners involved in direct treatment, such as physicians, nurses, addiction specialists, and counselors, who offer medical, therapeutic, and harm-reduction support for patients. Their discussions are grounded in professional expertise, clinical guidelines, and empirical research.

***QUERY STARTS HERE***
{query}
***QUERY ENDS HERE***

CHOOSE ONE AND WRAP YOUR ANSWER IN TRIPLE BRACKETS (e.g., [[[*YOUR ANSWER*]]])
- Patient
- Caregiver
- Practitioner

With your answer, please justify your reasoning (1-3 sentence).
\end{lstlisting}
\end{promptbox}

\subsection*{Prompt for Evaluating Content Preservation}
\begin{promptbox}
\begin{lstlisting}
How well does the query keep the core meaning of the original high-level question?

Original High-Level Question:
"{original_query}"

***QUERY STARTS HERE***
{detailed_query}
***QUERY ENDS HERE***

How well do they match in meaning? CHOOSE ONE AND WRAP YOUR ANSWER IN TRIPLE BRACKETS (e.g., [[[*YOUR ANSWER*]]])
- Completely dissimilar 
- Not equivalent, but share some details 
- Roughly equivalent 
- Mostly equivalent 
- Completely equivalent

With your answer, please justify your reasoning (1-3 sentence).
\end{lstlisting}
\end{promptbox}

\subsection{Human Annotation}
\label{sec:appendix-sim-human-eval}
\begin{figure*}[t]
\centering
\includegraphics[width=2\columnwidth]{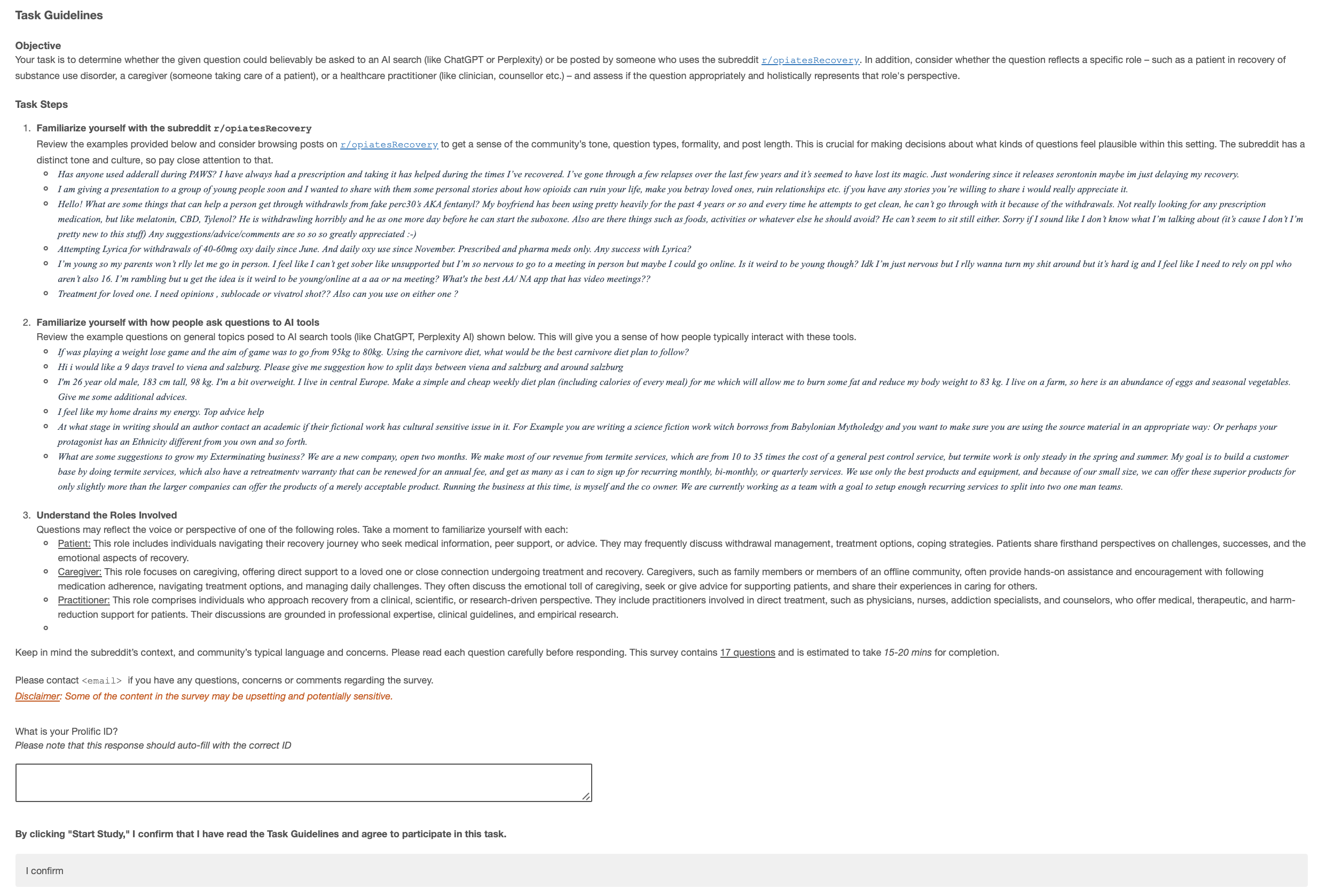}
\caption{Task Guidelines for human evaluation of believability of simulated role-based questions.}
\label{fig:appendix-task-guidelines} 
\end{figure*}

\begin{figure}[t]
\centering
\includegraphics[width=0.8\columnwidth]{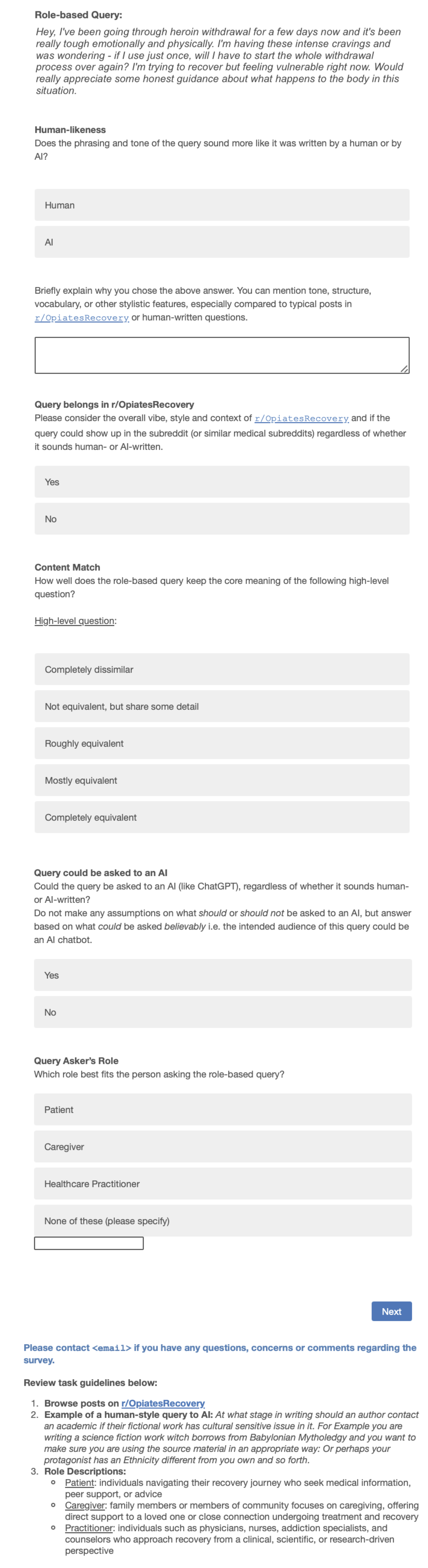}
\caption{User Interface for Human Annotations.}
\label{fig:appendix-ui} 
\end{figure}
As part of our evaluation setup, we validate GPT-4.1 judgments with $7{,}365$ human annotations on $491$ queries (Table \ref{tab:sim-eval-distribution}), with three annotators independently performing the same task as the GPT-4.1 judge. This section provides full details of the human evaluation process.

\paragraph{Evaluation Dataset}
The human evaluation dataset draws from three sources: 
(i) $31$ Reddit posts stratified across roles ($7$ each for patient, caregiver, and practitioner), using post-level role annotations and the extracted role-agnostic information-seeking questions described in \S\ref{sec:simulation}; 
(ii) $450$ simulated questions, obtained by sampling $30$ role-agnostic questions and their corresponding $3$ role-based simulated questions for each role, across \methodname and four prompting variants; and 
(iii) $10$ health- or clinical-related queries from the Wildchat dataset \cite{zhao2024wildchat}, identified using task and domain labels from \citet{mireshghallah2024trust}. While OUD-specific Wildchat queries would have been preferable, none were present, likely due to topic sensitivity and the public nature of the dataset.

\paragraph{Annotator Selection}
While moderators or community members of \texttt{r/OpiatesRecovery} familiar with AI tools would be the ideal evaluators, the sensitivity of the domain makes this infeasible. Instead, we recruit annotators via Prolific, filtered to (i) U.S. residents (where OUD discussions are most contextually relevant), (ii) active Reddit users, and (iii) English as their primary language.

Annotators first review detailed guidelines, including sample \texttt{r/OpiatesRecovery} posts to understand tone and norms, Wildchat examples to contextualize chatbot queries, and an overview of the role taxonomy and evaluation criteria (see Figures \ref{fig:appendix-task-guidelines} and \ref{fig:appendix-ui}). To ensure task comprehension, we include four qualification questions developed and validated in a pilot study by two expert (author) annotators; only participants who answer all correctly can proceed.

\paragraph{Task Setup}
Annotators are shown pairs of questions---a role-agnostic information-seeking question and its corresponding role-based simulated question---and asked to evaluate them across five metrics mentioned in \S\ref{sec:appendix-sim-eval-metrics}. We show task guidelines and user interface in Figures \ref{fig:appendix-task-guidelines} and \ref{fig:appendix-ui}. 
Each annotation task consists of $17$ question pairs sampled randomly from the full pool (Table \ref{tab:sim-eval-distribution}). A pilot study established an average completion time of $25$ minutes, and compensation is set to align with an hourly rate of USD12. In the full study, annotator completion times ranged from $10{-}70$ minutes.

Each question pair was independently rated by three annotators. To assess reliability, we compute inter-annotator agreement using Gwet's AC1, which is more robust than Fleiss' $\kappa$ under skewed label distributions \cite{ohyama2021statistical, wongpakaran2013comparison}. This is appropriate for our setting, as many simulated questions are expected to exhibit the target property, leading to imbalanced distributions. Metric-wise agreement scores are in Table \ref{tab:llm-judge-validation}.

\paragraph{Subjectivity in Human Annotations}
Because judgments of \textit{human-likeness} are inherently subjective and prior work shows that humans are not always reliable judges of AI-generated text \cite{jakesch2023human}, we further familiarized annotators with examples of both human- and AI-written text and required free-text rationales for their labels. These steps improved consistency but could not fully eliminate subjectivity in evaluations.

\begin{table}
\centering
\small
\begin{tabular}{@{}lcc@{}}
\toprule
\textbf{Metric} & \textbf{Gwet's AC1} & \textbf{Accuracy} \\ \midrule
Human Likeness & $0.63$ & $0.70$ \\
Context Plausibility & {$0.65$} & $0.78$ \\
Conversational AI Plausibility & $0.88$ & $0.90$ \\
Role Faithfulness & $0.99$ & $0.89$ \\ \bottomrule
\end{tabular}
\caption{
Inter-annotator agreement between human annotators (Gwet's AC1), and agreement between majority vote and GPT-4.1 rater (Accuracy).
}
\label{tab:llm-judge-validation}
\end{table}

\begin{table}
\centering
\small
\begin{tabular}{@{}lcc@{}}
\toprule
\textbf{Method} & \textbf{No. of role mentions} & \textbf{Proportion (\%)} \\ \midrule
\textsc{R}  & $1883$ & $66.5$ \\
\textsc{RG} & $2050$ & $72.4$ \\
\textsc{RGB} & $941$  & $33.2$ \\
\textsc{RGBE}  & $501$  & $17.7$ \\
\methodname & $0$ & $0$ \\ \bottomrule
\end{tabular}
\caption{Number of role-based questions with explicit role mentions (such as ``I am a patient \ldots''), which may inflate the \textit{role faithfulness} judgments.}
\label{tab:explicit-role-mentions}
\end{table}

\begin{table}
\centering
\small
\resizebox{\linewidth}{!}{\begin{tabular}{@{}lcccc@{}}
\toprule
 & \textbf{Patient} & \textbf{Caregiver} & \textbf{Practitioner} & \textbf{Overall} \\ \midrule
Human Likeness & $100.00 $ & $100.00$ & $78.50$ & $92.83$ \\
Context Plausibility & $99.68$ & $99.47$ & $ 85.06$ & 
$94.74$ \\
Interaction Plausibility & $80.51$ & $90.68$ & $97.25$ & $ 89.48$ \\
Role Faithfulness & $98.83$ & $99.89$ & $99.36$ & $99.36$\\
Content Preservation & $99.05$ & $99.89$ & $94.60$ & $97.84$ \\ \bottomrule
\end{tabular}}
\caption{Evaluating believability of \methodname generated questions (\S\ref{sec:sim-eval-results})}
\label{tab:corus-eval}
\end{table}

\section{Experimental Details}
\label{sec:appendix-exp-details}

\subsection{Model Details}
Table \ref{tab:appendix-model-details} lists the identifiers of all models used in our methodology and experiments.
Models were accessed via their organization's official APIs, except for Llama and OpenBioLLM-70B, which we accessed via the Nebius API.
For LLM judges, we set the temperature to $0$ to minimize output randomness. For the evaluated LLMs (\S\ref{sec:results-audit}), we used a temperature of $0.6$ and max\_tokens of $512$, following prior work \cite{chandra2024lived, agarwal2024medhalu}.

\begin{table}
\centering
\small
\begin{tabular}{@{}ll@{}}
\toprule
\textbf{Model Name}  
& \textbf{Identifier} \\ \midrule
GPT-4.1  
& \texttt{gpt-4.1-2025-04-14}  \\
GPT-4o  
& \texttt{gpt-4o-2024-08-06}  \\
GPT-4o-mini   
& \texttt{gpt-4o-mini-2024-07-18} \\
Claude-3-Haiku
& \texttt{claude-3-haiku-20240307} \\
Claude-3.7-Sonnet   
& \texttt{claude-3-7-sonnet-20250219}  \\
Gemini-2.5-Flash
& \texttt{gemini-2.5-flash}  \\
Llama-3.1-8B  
& \texttt{Llama-3.1-8B-Instruct}\\
Llama-3.1-70B
& \texttt{Meta-Llama-3.1-70B-Instruct}  \\
OpenBioLLM-70B 
& \texttt{aaditya/Llama3-OpenBioLLM-70B}  \\
\bottomrule
\end{tabular}
\caption{List of models used in our methodology and experiments, and their official identifiers.}
\label{tab:appendix-model-details}
\end{table}

\subsection{Response Evaluation}

\paragraph{Validating Knowledge and Support Classifiers}
Prior work has used the classifiers of \citet{choi2020ten} to study peer support \cite {balsamo2023pursuit}, conversation types \cite{zhou2021role}, and framing strategies \cite{mittal2025online}.
We use them to measure knowledge and support, and validate these classifiers for our setting to ensure they reliably capture these dimensions in our domain. 
Two authors independently annotated the $20$ highest- and $20$ lowest-scoring responses per dimension ($80$ total) with a binary label (high or low), after reaching a shared understanding of the criteria. 
We observe high inter-annotator agreement (Cohen’s $\kappa$ $0.7$) and strong alignment with classifier scores (accuracy $0.9$). 

\paragraph{Identifying Abstention in Responses}
We define abstention as cases where the model refuses to answer a query. To measure abstention rates, we detect such responses using a phrase-matching method with heuristics adapted from prior work \cite{arditi2024refusal, vasisht2024knowledge}. The method expands contractions (e.g., `can't' to `cannot'), looks for refusal markers (e.g., `I am unable', `I cannot support'), and applies a length-based heuristic to avoid false positives when the response later shifts into providing an answer. To validate, we manually checked $50$ flagged responses and found the method to be reasonably reliable ($0.73$ F-1 score).
\begin{table}
\centering
\small
\begin{tabular}{@{} m{\columnwidth}@{}}
\toprule
\textbf{Example Refused Query} \\ \midrule
My son swears he's been clean for 2 weeks but his test still showed fentanyl. Is that even possible? I've heard it can stay in fat cells. I want to believe him but I've been let down before... just trying to understand if this could be true or if he's using again. \\ \cmidrule{1-1}
My mom has been on pain meds for her back for almost a year now and I'm worried she needs more and more to get the same relief. Is there anything we can do to slow down how fast she's building tolerance? Her doctor says she'll need these for a long time and I'm scared about where this is heading. \\ \cmidrule{1-1}
my son tried oxy last week and now he keeps talking about how amazing it felt. i'm so worried he's going to get hooked. how do i help him stop thinking about that high? he's never been into drugs before this. \\ \cmidrule{1-1}
my son was using oxy for 2 days (around 150mg total) and has been clean for 6 days now. he's having minimal withdrawal symptoms and has a fast metabolism. is it safe for him to get the vivitrol shot now or should we wait longer? really worried about making the wrong call here \\ \bottomrule
\end{tabular}
\caption{Examples of queries which are refused}
\label{tab:appendix-refused-query-examples}
\end{table}

\section{Extended Results}
\label{sec:appendix-results}

\begin{figure*}[t]
    \centering
    \begin{subfigure}{0.66\columnwidth}
        \includegraphics[width=\columnwidth]{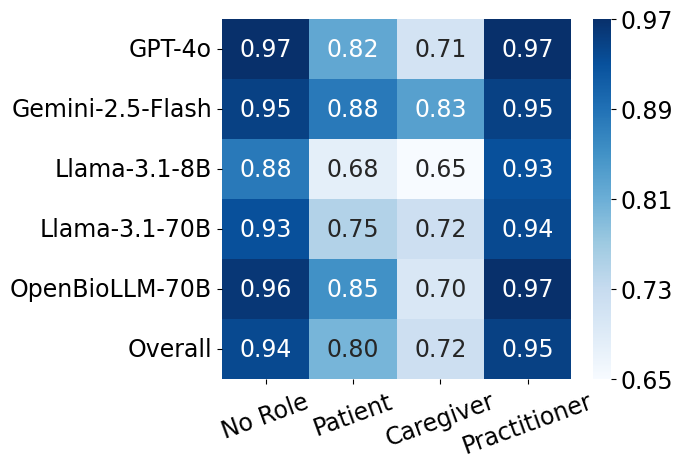}
        \caption{Knowledge Scores}
        \label{fig:heatmap-knowledge}
    \end{subfigure}
    \begin{subfigure}{0.66\columnwidth}
        \includegraphics[width=\columnwidth]{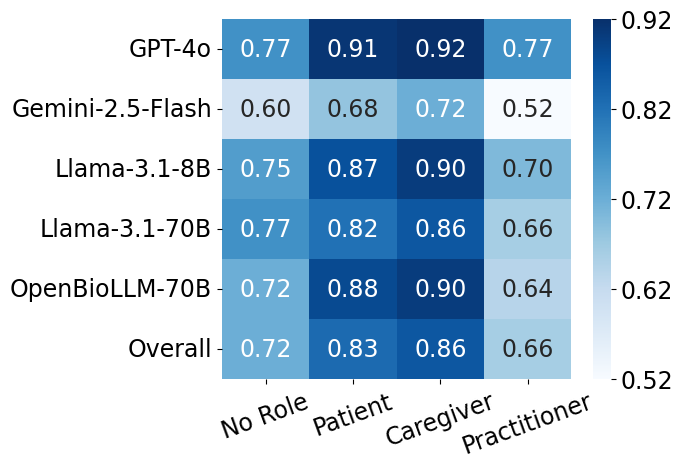}
        \caption{Support Scores}
        \label{fig:heatmap-support}
    \end{subfigure}
    \begin{subfigure}{0.66\columnwidth}
        \includegraphics[width=\columnwidth]{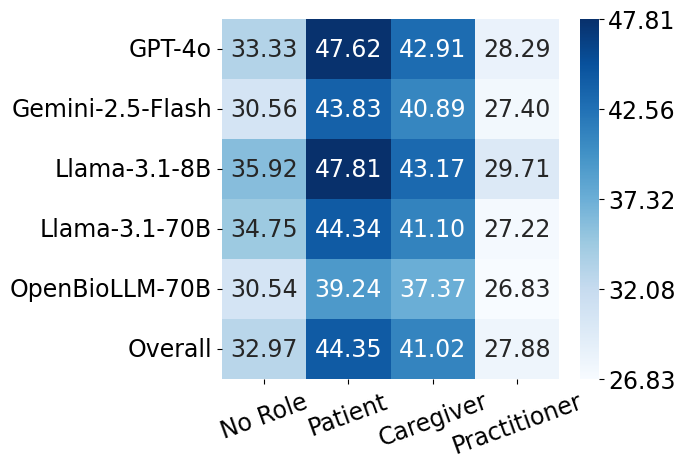}
        \caption{Readability Scores}
        \label{fig:heatmap-readability}
    \end{subfigure}
    \caption{Scores of model responses across (a) Knowledge, (b) Support, and (c) Readability when given a query without context (\texttt{No Role}) and with role-based context of \texttt{Patient}, \texttt{Caregiver} and \texttt{Practitioner}. }
    \label{fig:llm-eval-context}
\end{figure*}

\begin{figure*}[t]
\centering
\includegraphics[width=2\columnwidth]{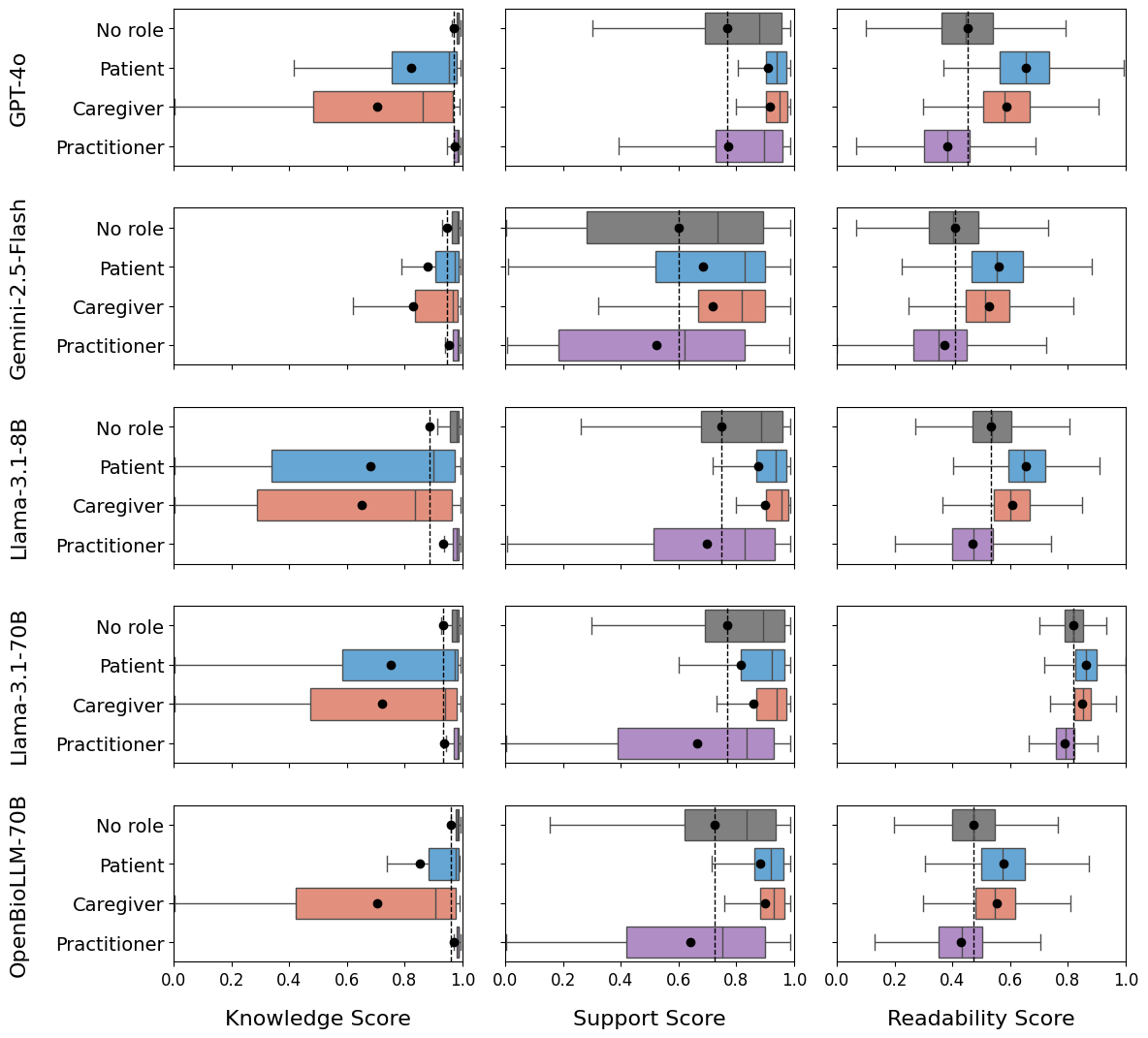} 
\caption{Distribution of Knowledge, Support and Readability scores of model responses across role-agnostic and role-based questions.}
\label{fig:box-plot-llm-eval}
\end{figure*}

\subsection{Comparison between \methodname and Real-User Queries}
As compared to \textsc{CoRUS}-generated queries, although Reddit queries are more \textit{human-like} ($93\%$ vs. $100\%$) and \textit{contextual plausibility} ($94\%$ vs. $98\%$), the effect sizes are small to moderate ($h = 0.4$ and $h = 0.2$, respectively), indicating that these differences are not practically large.
For \textit{human-likeness}, \methodname and WildChat perform similarly, with no significant difference, confirming that the generated queries are judged as human-like.
For \textit{interaction plausibility}, \methodname performs strongly ($89\%$), surpassing Reddit ($59\%$, $h = 0.7$, large effect) and approaching WildChat ($97\%$, $h = 0.3$, small effect).
Finally, \methodname achieves high \textit{role faithfulness} ($95\%$), only slightly less than Reddit ($98\%$, $h = 0.2$, small effect).
Importantly, \methodname avoids the limitations of Reddit (low interaction plausibility) and WildChat (no OUD- or role-specific framing), making it the only setting that performs consistently well across all metrics.

\begin{table*}
\centering
\scriptsize
\begin{tabular}{@{}
m{0.4\columnwidth} m{1.5\columnwidth}@{}}      
\toprule
Role-based Question 
& GPT-4o Response \\ \midrule
\colorbox{ptcolor!50}{Patient}: these cravings are killing me man. my whole body feels like it's screaming for oxy. been clean for 9 days but i'm literally shaking sometimes. what do people do to get through this? i'm trying to stay strong but it's so hard 
& I'm really sorry to hear that you're going through this. It's incredibly difficult to deal with cravings and withdrawal symptoms, but reaching out for support is a strong and important step. Here are some strategies that might help:\newline 1. Reach Out for Support: Contact a support group like Narcotics Anonymous (NA) or SMART Recovery. Connecting with others who understand what you're going through can provide immense support.\newline 2. Counseling or Therapy: Consider talking to a counselor or therapist who specializes in addiction. They can offer guidance and coping strategies tailored to your needs.\newline 3. Medical Support: Consult with a healthcare professional about your symptoms. They might prescribe medications that can help manage withdrawal symptoms and cravings.\newline 4. Stay Busy: Engage in activities that keep your mind and body occupied. Exercise, hobbies, or even simple tasks can help distract you from cravings.\newline 5. Mindfulness and Relaxation: Practices such as meditation, deep breathing exercises, or yoga can help reduce stress and improve your ability to cope with cravings.\newline  6. Healthy Lifestyle: Eating well, staying hydrated, and getting enough sleep can help your body recover and improve your mood and energy levels.\newline 7. Set Short-Term Goals: Focus on getting through one day or even one hour at a time. Celebrate small victories to keep yourself motivated.\newline  8. Avoid Triggers: Identify people, places, or situations that trigger cravings and try to avoid them as much as possible.\newline  Remember, it's okay to ask for help, and seeking professional support can be crucial. You're not alone in this, and there are people and resources available to support you on your journey to recovery. Stay strong and take it one step at a time.\\ \cmidrule{1-2}
\colorbox{cgcolor!50}{Caregiver}: my son is 2 weeks clean from fentanyl and he's having these awful full body cravings that make him so restless. i don't know how to help him through this. what can i do to make it easier for him? he's trying so hard but i can see how much he's suffering
& I'm sorry to hear that your son is going through such a difficult time, and it's understandable that you want to support him in any way you can. Here are some ways you might be able to help him through this challenging period:\newline 1. Encourage Professional Support: If he isn't already, encourage your son to work with healthcare professionals who specialize in addiction treatment. They can provide medical support, counseling, and possibly medication-assisted treatment to help manage cravings and withdrawal symptoms.\newline 2. Create a Supportive Environment: Make your home a safe and supportive space. Remove any triggers that might remind him of using, and encourage a routine that promotes healthy habits.\newline 3. Encourage Healthy Coping Strategies: Help him find healthy ways to cope with cravings, such as engaging in physical activity, practicing mindfulness or meditation, or pursuing hobbies that distract and occupy his mind.\newline 4. Stay Connected: Regularly check in with him and offer your emotional support. Let him know that you are there for him and that you believe in his ability to overcome this challenge.\newline 5. Provide Resources: Share information about local support groups or online communities where he can connect with others who are going through similar experiences. Sometimes, talking to others who understand can be incredibly reassuring.\newline 6. Educate Yourself: Learn as much as you can about addiction and recovery. Understanding what he's going through can help you provide better support and empathy.\newline 7. Encourage Self-care: Remind him to take care of himself by eating well, staying hydrated, getting enough sleep, and practicing relaxation techniques.\newline 8. Be Patient and Non-judgmental: Recovery is a long and challenging process with ups and downs. Be patient with him and avoid any judgments or criticisms that could make him feel worse.\newline 9. Seek Support for Yourself: Supporting a loved one through recovery can be emotionally taxing. Consider joining a support group for family members of people with addiction, such as Nar-Anon or Al-Anon, to find support and share experiences with others.\newline Remember, while you can provide support and encouragement, the decision to stay clean and seek help ultimately lies with him. It's important to take care of yourself, too, so that you can continue to be a source of strength for your son. \\ \cmidrule{1-2}
\colorbox{prcolor!50}{Practitioner}:  I've got several patients struggling with these intense full body cravings while getting off oxycodone. The standard meds help somewhat, but they're still really suffering. What's working for other clinicians? Need some practical strategies beyond just clonidine and comfort meds that I can recommend tomorrow in clinic.
& Managing intense cravings and withdrawal symptoms in patients tapering off oxycodone can be challenging. While clonidine and comfort medications are a standard part of treatment, there are several additional strategies that clinicians have found helpful:\newline 1. Medication-Assisted Treatment (MAT): Consider using medications like buprenorphine/naloxone (Suboxone) or methadone as part of a comprehensive treatment plan. These medications can help reduce cravings and withdrawal symptoms significantly.\newline 2. Adjunctive Medications:\newline   - Gabapentin or Pregabalin: These can help with neuropathic pain and may have an anxiolytic effect, which can be beneficial for withdrawal symptoms.\newline   - Tizanidine or Baclofen: Muscle relaxants that can help with muscle cramps and spasms.\newline   - Lofexidine: An alternative to clonidine, it is specifically approved for opioid withdrawal and may cause less hypotension.\newline 3. Behavioral Therapies:\newline   - Cognitive Behavioral Therapy (CBT): Can help patients develop coping strategies to manage cravings and triggers.\newline   - Motivational Interviewing (MI): Helps to enhance the patient's motivation to change and adhere to the treatment plan.\newline 4. Support Groups and Counseling:\newline   - Encourage participation in support groups such as Narcotics Anonymous (NA) or other peer support programs.\newline   - Individual or group counseling can provide emotional support and help patients develop coping mechanisms.\newline 5. Lifestyle Modifications:\newline   - Encourage regular exercise, which can help reduce stress and improve mood.\newline   - Promote a healthy diet and adequate hydration to support overall well-being.\newline   - Mindfulness practices, such as meditation or yoga, can help manage stress and cravings.\newline 6. Acupuncture and Acupressure: Some patients find relief from withdrawal symptoms through these alternative therapies.\newline 7. Transcranial Magnetic Stimulation (TMS): Though still under investigation, TMS has shown promise in reducing cravings and withdrawal symptoms in some studies.\newline 8. Close Monitoring and Follow-up: Regular follow-up appointments to monitor progress and adjust the treatment plan as needed are crucial.\newline It's important to tailor the approach to each individual patient, as responses to treatments can vary. Combining pharmacological and non-pharmacological strategies often yields the best results.\\ \bottomrule
\end{tabular}
\caption{GPT-4o responses to role-based queries, corresponding to the role-agnostic query \textit{`How do individuals deal with intense full body cravings during recovery from opiates?'}}
\label{tab:gpt-responses}
\end{table*}

\begin{table*}
\centering
\scriptsize
\begin{tabular}{@{}
m{0.4\columnwidth} m{1.5\columnwidth}@{}} 
\toprule
Role-based Question 
& Gemini-2.5-Flash Response \\ \midrule
\colorbox{ptcolor!50}{Patient}: these cravings are killing me man. my whole body feels like it's screaming for oxy. been clean for 9 days but i'm literally shaking sometimes. what do people do to get through this? i'm trying to stay strong but it's so hard 
& It sounds like you're going through an incredibly tough time right now, and it takes immense strength to reach out and talk about what you're experiencing. The fact that you've made it 9 days clean is a huge accomplishment, especially when your body is screaming for something it's used to. Those physical and mental cravings, the shaking – that's your body adjusting, and it's a brutal process. Please know that what you're feeling is real and valid, and it's a testament to your commitment to getting clean.\newline It's completely understandable that you're finding it hard; this is one of the hardest things a person can go through. You're not alone in feeling this way. Many people have walked this path, and while everyone's journey is unique, there are common strategies and supports that have helped others get through these intense periods.\newline Here's what people often do to get through the kind of cravings and withdrawal you're describing:\newline 1. Immediate Distraction \& Ride the Wave:
 The 15-Minute Rule: Cravings often come in waves and can peak and then subside. Tell yourself, I just need to get through the next 15 minutes. Set a timer. During that time, do something, anything, to distract yourself.
 Engage Your Senses:
 Cold Shower/Ice: Dunking your face in ice water or taking a cold shower can be a shock to the system that resets your focus. Holding ice cubes in your hands can also work.
 Strong Smells: Sniffing something strong like peppermint oil, rubbing alcohol, or even a strong cleaning product (briefly and safely) can break the thought cycle.
 Intense Flavors: Chew strong gum, suck on a sour candy, or eat something spicy.
 Physical Activity (if safe and able): Even a short walk, some jumping jacks, or stretching can help release endorphins and shift your focus.
 Engage Your Mind: Do a puzzle, read a book, watch a captivating show, play a video game, listen to a podcast, or call someone. The goal is to occupy your mental space.\newline 2. Lean on Your Support System:
 Talk to Someone: This is huge. Call a trusted friend, family member, sponsor, or therapist. \\ \cmidrule{1-2}
\colorbox{cgcolor!50}{Caregiver}: my son is 2 weeks clean from fentanyl and he's having these awful full body cravings that make him so restless. i don't know how to help him through this. what can i do to make it easier for him? he's trying so hard but i can see how much he's suffering
& It's incredibly brave of your son to be two weeks clean from fentanyl, and it's completely understandable that you're feeling helpless and worried seeing him suffer. What he's experiencing are likely post-acute withdrawal symptoms (PAWS), which can be very challenging and last for weeks or even months after the initial acute withdrawal. The restlessness and full-body cravings are a common part of this.\newline Here's what you can do to help make it easier for him, focusing on both practical support and emotional understanding:\newline 1. Acknowledge and Validate His Pain:\newline  Tell him you see his struggle: I see how hard you're fighting this, and I know how much you're suffering right now. It breaks my heart to see you in so much pain, but I'm so incredibly proud of your strength.
 Reassure him it's normal: What you're feeling is a normal part of getting clean, even after two weeks. Your body and brain are still adjusting, and it's not a sign of weakness.
 Avoid judgment: Don't say things like, Just push through it, or You should be feeling better by now. This can make him feel more isolated and like he's failing.\newline 2. Provide a Calm and Supportive Environment:\newline  Reduce stress: Try to minimize any additional stressors in the household. Keep things calm and predictable.
 Quiet space: Ensure he has a quiet, comfortable space where he can rest or try to relax without too much stimulation.
 Be present: Sometimes just having you nearby, even if you're not talking, can be comforting.\newline 3. Address the Physical Discomfort (within reason):\newline  Warm baths/showers: These can be incredibly soothing for restless legs and body aches. Suggest Epsom salt baths if he's open to it.
 Gentle massage: If he's comfortable, offer a gentle foot, hand, or back massage. This can help with the restless feeling.
 Light exercise (if tolerated): Even a short walk around the block, some stretching, or gentle yoga can help release some of the restless energy and promote endorphins. Don't push it, though.
 Hydration and Nutrition: Make sure he's staying well-hydrated with water, herbal teas, or broths \\ \cmidrule{1-2}
\colorbox{prcolor!50}{Practitioner}: I've got several patients struggling with these intense full body cravings while getting off oxycodone. The standard meds help somewhat, but they're still really suffering. What's working for other clinicians? Need some practical strategies beyond just clonidine and comfort meds that I can recommend tomorrow in clinic.
& This is a challenging but important area, and it's great you're looking for more effective strategies for your patients. The full body craving sensation during opioid withdrawal, often described as a deep, pervasive ache or an intense internal restlessness, is particularly distressing and can significantly contribute to relapse. While clonidine and comfort meds are foundational, they often don't fully address this specific symptom.\newline Here are some practical strategies and medications that clinicians are finding helpful, beyond the standard, that you can consider for your patients tomorrow:\newline  I. Optimizing Existing Strategies \& Addressing Gaps:\newline  Buprenorphine Microdosing Low-Dose Induction:
 Why it helps: Even very low doses of buprenorphine (e.g., 0.5mg, 1mg, 2mg) can significantly alleviate opioid withdrawal symptoms, including the full-body cravings, by partially activating opioid receptors. This is often more effective than clonidine alone.
 Practical Tip: If patients are resistant to full MAT or you're concerned about precipitated withdrawal (though less likely with very low doses), consider a comfort dose of buprenorphine, even just for a few days, to bridge the acute withdrawal phase. Some clinicians will prescribe a few doses to be taken only if symptoms are severe.
 For patients who can't access full MAT right away: This can be a life-saver to get them through the worst of the physical symptoms while waiting for an MAT appointment.\newline  Gabapentin/Pregabalin Optimization:
 Why it helps: These meds are excellent for neuropathic pain and restless legs, but they also have an anxiolytic and sedative effect that can indirectly reduce the intensity of the body cravings and general discomfort. They can also help with sleep, which is often severely disrupted.
 Practical Tip: Ensure doses are adequate and titrated up quickly if tolerated. For gabapentin, consider q8h dosing (or even q6h in acute withdrawal) as its half-life is relatively short. Max out at 1800-3600mg/day as tolerated. For pregabalin, 150-600mg/day in divided doses. Be mindful of potential for misuse, especially with pregabalin.\\ \bottomrule
\end{tabular}
\caption{Gemini-2.5-Flash responses to role-based queries, corresponding to the role-agnostic query \textit{`How do individuals deal with intense full body cravings during recovery from opiates?'}}
\label{tab:gemini-responses}
\end{table*}

\begin{table*}
\centering
\scriptsize
\begin{tabular}{@{}
m{0.4\columnwidth} m{1.5\columnwidth}@{}} 
\toprule
Role-based Question 
& Llama-3.1-8B Response \\ \midrule
\colorbox{ptcolor!50}{Patient}: these cravings are killing me man. my whole body feels like it's screaming for oxy. been clean for 9 days but i'm literally shaking sometimes. what do people do to get through this? i'm trying to stay strong but it's so hard 
& I'm so sorry to hear that you're going through this. Withdrawal symptoms can be overwhelming, especially when your body starts to detox. It's great that you've made it 9 days clean, though - that's a huge accomplishment.I'm so sorry to hear that you're going through this. Withdrawal symptoms can be overwhelming, especially when your body starts to detox. It's great that you've made it 9 days clean, though - that's a huge accomplishment.\newline To get through this, here are some strategies that might help:\newline 1. Stay hydrated: Drink plenty of water, and consider adding electrolyte-rich beverages like coconut water or sports drinks to help replace lost salts.
2. Get moving: Engage in some light physical activity, like a short walk or some yoga stretches. Exercise can help reduce anxiety and cravings.
3. Mindfulness and meditation: Practice deep breathing exercises, meditation, or guided relaxation to calm your mind and body. You can find plenty of free resources online, like Headspace or Calm.
4. Support system: Reach out to friends, family, or a support group for emotional support. Sharing your struggles with others who understand can be incredibly helpful.
5. Healthy distractions: Engage in activities you enjoy, like reading, drawing, or listening to music. This can help take your mind off cravings and withdrawal symptoms.
6. Professional help: Consider consulting a doctor or a therapist who specializes in addiction. They can provide guidance on managing withdrawal symptoms and offer additional support.
7. Supplements: Certain supplements like vitamin B, magnesium, or omega-3 fatty acids might help alleviate withdrawal symptoms. However, always consult a doctor before adding any new supplements to your routine.
8. Sleep: Prioritize getting enough sleep each night. Lack of sleep can exacerbate withdrawal symptoms, so aim for 7-8 hours of sleep per night.
9. Reward system: Celebrate your small victories, like reaching a milestone day or completing a challenging task. This can help motivate you to stay on track.
10. Be kind to yourself: Remember that relapse is a part of the recovery process, and it doesn't define your worth as a person. If you do experience a setback, don't be too hard on yourself - instead, focus on getting back on track as soon as possible.\newline Remember, the first 30 days are often the most challenging, but with time and support, you can overcome your cravings and achieve long-term sobriety.\newline If you're feeling overwhelmed or need immediate support, consider reaching out to:\newline National Helpline: 1-800-662-HELP (4357) Substance Abuse and Mental Health Services Administration (SAMHSA)\\ \cmidrule{1-2}
\colorbox{cgcolor!50}{Caregiver}: my son is 2 weeks clean from fentanyl and he's having these awful full body cravings that make him so restless. i don't know how to help him through this. what can i do to make it easier for him? he's trying so hard but i can see how much he's suffering
& I'm glad you're seeking help and support for your son during this challenging time. Two weeks into recovery from fentanyl can be a difficult period, and full-body cravings are common during this phase. Here are some suggestions that may help alleviate your son's discomfort and make the recovery process easier:\newline 1. Stay calm and patient: It's essential to remain calm and patient when dealing with your son's cravings. Avoid getting frustrated or anxious, as this can escalate the situation.
2. Encourage physical activity: Regular exercise can help reduce cravings and alleviate restlessness. Suggest activities like walking, jogging, yoga, or even dancing to your son. Exercise can also help release endorphins, which can improve mood.
3. Offer healthy distractions: Engage your son in activities that require his full attention, such as puzzles, drawing, or playing a game. This can help distract him from cravings and provide a sense of accomplishment.
4. Provide a comfortable environment: Create a peaceful and comfortable space for your son to relax. Ensure the room is quiet, dark, and at a comfortable temperature. You can also use aromatherapy or calming music to promote relaxation.
5. Support healthy sleep habits: Encourage your son to establish a regular sleep schedule and create a cozy sleep environment. A well-rested individual is better equipped to manage cravings.
6. Encourage hydration and nutrition: Adequate hydration and a balanced diet can help alleviate cravings and reduce withdrawal symptoms. Suggest healthy snacks and encourage your son to drink plenty of water.
7. Medication-assisted treatment (MAT): If your son is not already on MAT, discuss this option with his healthcare provider. Medications like buprenorphine or methadone can help manage withdrawal symptoms and cravings.
8. Counseling and therapy: Encourage your son to attend individual or group counseling sessions. A therapist can help him develop coping strategies, address underlying issues, and provide emotional support.
9. Support groups: Consider attending support groups with your son, such as Narcotics Anonymous (NA) or SMART Recovery. Sharing experiences and connecting with others who are going through similar struggles can be incredibly helpful.
10. Stay connected with his healthcare provider: Regular check-ins with his healthcare provider can help monitor progress, adjust treatment plans, and address any concerns or issues that arise.
11. Educate yourself: Learn about fentanyl withdrawal, recovery, and relapse prevention. Understanding the process and being aware of potential triggers can help you \\ \cmidrule{1-2}
\colorbox{prcolor!50}{Practitioner}: I've got several patients struggling with these intense full body cravings while getting off oxycodone. The standard meds help somewhat, but they're still really suffering. What's working for other clinicians? Need some practical strategies beyond just clonidine and comfort meds that I can recommend tomorrow in clinic.
& Managing intense cravings for patients withdrawing from oxycodone can be challenging. Here are some additional strategies that clinicians have found helpful:\newline 1. Mindfulness and meditation: Encourage patients to practice mindfulness techniques, such as deep breathing, progressive muscle relaxation, or guided meditation. This can help them focus on the present moment and reduce cravings.
2. Behavioral activation: Help patients engage in activities that bring them joy and fulfillment, such as hobbies, exercise, or spending time with loved ones. This can help distract them from cravings and improve mood.
3. Cognitive-behavioral therapy (CBT): CBT can help patients identify and challenge negative thought patterns and behaviors associated with cravings. A therapist can work with the patient to develop coping skills and strategies to manage cravings.
4. Acupuncture: Some studies suggest that acupuncture may be effective in reducing withdrawal symptoms, including cravings, in individuals withdrawing from opioids.
5. Topiramate: This medication has been shown to reduce cravings and improve outcomes in individuals with opioid use disorder. It works by reducing the release of dopamine and glutamate in the brain.
6. Naltrexone implants: For patients with a history of relapse, naltrexone implants can be an effective option. These implants release naltrexone, an opioid antagonist, over several months, helping to reduce cravings and prevent relapse.
7. Buprenorphine with naloxone: Adding naloxone to buprenorphine can help deter patients from injecting or snorting the medication, reducing the risk of overdose and misuse.
8. Kratom: Some clinicians have reported success with kratom in reducing cravings and withdrawal symptoms. However, its use is not FDA-approved, and more research is needed to fully understand its effects.
9. Exercise and physical activity: Regular exercise has been shown to reduce cravings and improve mood in individuals with opioid use disorder.
10. Support groups: Encourage patients to participate in support groups, such as Narcotics Anonymous (NA) or SMART Recovery, to connect with others who are going through similar experiences.
11. Cannabidiol (CBD): Some studies suggest that CBD may be effective in reducing anxiety and stress associated with cravings.
12. Pharmacological agents: Other medications that have been used to treat cravings include disulfiram, acamprosate, and gabapentin. However, more research is needed to fully understand their effectiveness.\\ \bottomrule
\end{tabular}
\caption{Llama-3.1-8B responses to role-based queries, corresponding to the role-agnostic query \textit{`How do individuals deal with intense full body cravings during recovery from opiates?'}}
\label{tab:llama-8b-responses}
\end{table*}

\begin{table*}
\centering
\scriptsize
\begin{tabular}{@{}
m{0.4\columnwidth} m{1.5\columnwidth}@{}}      
\toprule
Role-based Question 
& Llama-3.1-70B  Response \\ \midrule
\colorbox{ptcolor!50}{Patient}: these cravings are killing me man. my whole body feels like it's screaming for oxy. been clean for 9 days but i'm literally shaking sometimes. what do people do to get through this? i'm trying to stay strong but it's so hard 
& Overcoming Opioid Cravings: Strategies for Success\newline Here are some strategies that people find helpful:\newline    Reach out for support: Talk to a trusted friend, family member, or sponsor about the cravings. Sharing the experience with someone who understands can help alleviate the emotional burden.\newline - Stay hydrated: Drink plenty of water and other fluids to help flush out the body and reduce withdrawal symptoms.\newline - Engage in physical activity: Exercise, such as walking or yoga, can help release endorphins, which can improve mood and reduce cravings.\newline - Practice relaxation techniques: Activities like meditation, deep breathing, or progressive muscle relaxation can help calm the mind and body.\newline - Use positive self-talk: Encourage yourself with positive affirmations, such as I am strong, I can get through this, or I am worthy of recovery.\newline - Seek professional help: Consider consulting a therapist or counselor who specializes in addiction recovery. They can provide guidance, support, and evidence-based treatments to help manage cravings and stay on track. \\ \cmidrule{1-2}
\colorbox{cgcolor!50}{Caregiver}: my son is 2 weeks clean from fentanyl and he's having these awful full body cravings that make him so restless. i don't know how to help him through this. what can i do to make it easier for him? he's trying so hard but i can see how much he's suffering
& I'm so glad you're supporting your son through this challenging time. Full-body cravings can be a particularly difficult symptom of opioid withdrawal, but there are some strategies that may help alleviate his discomfort. Please keep in mind that these suggestions are not a substitute for medical professional advice, and it's essential to consult with his treatment team for personalized guidance.\newline Here are some ideas to help your son cope with full-body cravings:\newline 1. Stay hydrated: Encourage your son to drink plenty of water, clear broths, or electrolyte-rich beverages like coconut water or sports drinks. Sometimes, dehydration can exacerbate withdrawal symptoms.
2. Warm or cold compresses: Applying a warm or cold compress to his body may help relax his muscles and reduce restlessness. You can use a heating pad, warm bath, or a cold pack wrapped in a towel.
3. Gentle exercise: Encourage your son to engage in light physical activity, like short walks, yoga, or stretching exercises. This can help release endorphins, which can help alleviate cravings.
4. Mindfulness techniques: Teach your son some mindfulness exercises, such as deep breathing, progressive muscle relaxation, or guided meditation. These can help him focus on the present moment and reduce anxiety.
5. Comfort measures: Offer physical comfort like hugs, holding hands, or gentle massages. Sometimes, human touch can be incredibly soothing.
6. Distract and redirect: Engage your son in activities that distract him from his cravings, such as watching a movie, playing a game, or listening to music.
7. Hot showers or baths: A warm shower or bath can be very relaxing and help reduce muscle tension.
8. Over-the-counter medications: Certain medications like ibuprofen or acetaminophen may help alleviate some withdrawal symptoms, but always consult with his treatment team before giving him any medication.
9. Caffeine and sugar: A small amount of caffeine or a sugary snack may help alleviate some symptoms, but be cautious not to overdo it, as this can lead to other problems.
10. Support groups: Encourage your son to attend support groups, such as Narcotics Anonymous (NA) or SMART Recovery, where he can connect with others who are going through similar experiences.
11. Medication-assisted treatment (MAT): If your son's treatment team has prescribed MAT, such as buprenorphine or methadone, ensure he's taking it as directed. \\ \cmidrule{1-2}
\colorbox{prcolor!50}{Practitioner}:  I've got several patients struggling with these intense full body cravings while getting off oxycodone. The standard meds help somewhat, but they're still really suffering. What's working for other clinicians? Need some practical strategies beyond just clonidine and comfort meds that I can recommend tomorrow in clinic.
& Here are some practical strategies that have shown promise in managing withdrawal symptoms:\newline 1.  Mindfulness and meditation: Encourage patients to practice mindfulness techniques, such as deep breathing, progressive muscle relaxation, or guided meditation. These can help reduce anxiety and stress, which can exacerbate cravings.
2.  Physical activity: Regular exercise, such as yoga or walking, can help reduce withdrawal symptoms and improve mood.
3.  Aromatherapy: Certain essential oils like lavender, chamomile, and bergamot have been shown to have a calming effect and may help reduce anxiety and cravings.
4.  Nutritional support: Ensure patients are getting adequate nutrition, including plenty of fruits, vegetables, whole grains, and lean protein sources. Consider recommending supplements like vitamin B complex, magnesium, and omega-3 fatty acids, which can help alleviate withdrawal symptoms.
5.  Cognitive-behavioral therapy (CBT): CBT can help patients identify and challenge negative thought patterns and behaviors that contribute to cravings.
6.  Acupuncture: This traditional Chinese medicine technique has been shown to reduce withdrawal symptoms and cravings in some studies.
7.  Massage therapy: Massage can help reduce stress, anxiety, and muscle tension, which can contribute to cravings.
8.  Sleep hygiene: Establish a consistent sleep schedule, avoid caffeine and electronics before bedtime, and create a relaxing bedtime routine to improve sleep quality.
9.  Social support: Encourage patients to connect with friends, family, or support groups to help them feel less isolated and more motivated to stay on track with their recovery.
10. Alternative medications: Consider prescribing alternative medications like gabapentin, pregabalin, or baclofen, which have been shown to be effective in reducing withdrawal symptoms and cravings in some patients.\newline These strategies can be used in conjunction with standard medications like clonidine and comfort meds to provide a more comprehensive treatment plan. \\ \bottomrule
\end{tabular}
\caption{Llama-3.1-70B responses to role-based queries, corresponding to the role-agnostic query \textit{`How do individuals deal with intense full body cravings during recovery from opiates?'}}
\label{tab:llama-70b-responses}
\end{table*}

\begin{table*}
\centering
\scriptsize
\begin{tabular}{@{}
m{0.4\columnwidth} m{1.5\columnwidth}@{}}      
\toprule
Role-based Question 
& OpenBioLLM-70B  Response \\ \midrule
\colorbox{ptcolor!50}{Patient}: these cravings are killing me man. my whole body feels like it's screaming for oxy. been clean for 9 days but i'm literally shaking sometimes. what do people do to get through this? i'm trying to stay strong but it's so hard 
& I'm not a doctor, but I can offer some general advice. It's important to remember that recovering from addiction takes time and it's normal to experience cravings and withdrawal symptoms. Here are a few things that may help you get through this difficult period:  1. Seek support: Reach out to a trusted friend, family member, or support group who can provide emotional support during this challenging time.  2. Stay active: Engaging in physical activities such as exercise or sports can help reduce cravings and improve overall well-being.  3. Find healthy distractions: Keep yourself occupied with activities that you enjoy, such as hobbies, creative pursuits, or spending time in nature.  4. Practice relaxation techniques: Deep breathing exercises, meditation, or yoga can help manage stress and anxiety related to cravings.  5. Seek professional help: Consider reaching out to a healthcare professional or addiction counselor who can provide guidance and support tailored to your specific situation.  Remember, recovery is a journey, and it's okay to ask for help along the way. \\ \cmidrule{1-2}
\colorbox{cgcolor!50}{Caregiver}: my son is 2 weeks clean from fentanyl and he's having these awful full body cravings that make him so restless. i don't know how to help him through this. what can i do to make it easier for him? he's trying so hard but i can see how much he's suffering
& I'm not a doctor, but I can offer some general information that may help. It's important to remember that only a healthcare professional can provide specific advice tailored to your son's situation.  Fentanyl is a potent opioid, and withdrawal symptoms can be severe. The duration and intensity of withdrawal vary among individuals. Here are some suggestions that may help alleviate your son's symptoms:  1. Seek professional help: Encourage your son to consult with a healthcare provider who specializes in addiction treatment. They can assess his condition, provide appropriate medical interventions, and recommend strategies to manage withdrawal symptoms.  2. Medication-assisted treatment (MAT): Depending on the severity of your son's addiction and withdrawal symptoms, a healthcare provider may prescribe medications such as methadone, buprenorphine, or naltrexone to assist in managing cravings and reducing withdrawal symptoms. These medications should only be used under medical supervision.  3. Support groups: Consider attending support groups like Narcotics Anonymous (NA) or other community-based programs that offer guidance and emotional support for individuals recovering from drug addiction. Sharing experiences and connecting with others who are going through similar challenges can be beneficial.  4. Healthy lifestyle habits: Encourage your son to maintain a healthy diet, exercise regularly, get enough sleep, and practice stress-reducing techniques such as deep breathing exercises, meditation, or yoga. These lifestyle changes can help improve overall well-being and reduce anxiety associated with withdrawal.  5. Create a supportive environment: Offer emotional support by being understanding and patient. Ensure a safe and comfortable living space for your son. Remove any triggers or reminders of drug use from his environment that might exacerbate cravings.  6. Distract and engage: Help your son find healthy distractions from cravings, such as engaging in hobbies, spending time with supportive friends or family members, or pursuing new interests. Keeping busy and focused on positive activities can make it easier to cope with withdrawal symptoms.  7. Be aware of potential complications: Monitor your son closely for any signs of severe symptoms or complications during withdrawal. If he experiences extreme pain, difficulty breathing, rapid heartbeat, or any other concerning symptoms, seek immediate medical attention.  Remember, overcoming addiction is a challenging journey, and relapse can occur. It's essential to approach recovery with compassion and understanding. Encourage your son to stay committed to his treatment plan and seek ongoing support.  The Answer is to consult with a healthcare provider who specializes in addiction treatment for personalized advice and guidance. \\ \cmidrule{1-2}
\colorbox{prcolor!50}{Practitioner}:  I've got several patients struggling with these intense full body cravings while getting off oxycodone. The standard meds help somewhat, but they're still really suffering. What's working for other clinicians? Need some practical strategies beyond just clonidine and comfort meds that I can recommend tomorrow in clinic.
& Managing withdrawal symptoms during opioid tapering can be challenging. While clonidine and comfort medications are commonly used, there are several additional strategies that can help alleviate these symptoms. Here are some practical strategies you can recommend to your patients:  1. Adjuvant medications: Consider adding other medications such as alpha-2 adrenergic agonists (e.g., guanfacine) or beta-blockers (e.g., propranolol) to help manage autonomic symptoms like sweating, anxiety, and tremors.  2. Gabapentin: This medication has shown promise in reducing opioid withdrawal symptoms, particularly anxiety, insomnia, and muscle pain. It can be started before the taper and continued through the withdrawal process.  3. Baclofen: This muscle relaxant can be effective in managing muscle spasms, cramps, and anxiety during opioid withdrawal. It may also have a role in reducing cravings and aiding sleep.  4. Clonidine patch: Instead of using oral clonidine, consider prescribing a clonidine transdermal patch. This provides a steady release of the medication over 7 days and may improve compliance.  5. Non-pharmacological interventions:    - Mindfulness-based stress reduction (MBSR) or meditation techniques can help patients cope with cravings and anxiety.    - Acupuncture has shown some benefit in reducing withdrawal symptoms and cravings.    - Physical activity, such as regular exercise or yoga, can aid in stress reduction and improve overall well-being.    - Adequate sleep hygiene and practicing good sleep habits are crucial during this period.  6. Nutritional support: Encourage patients to maintain a balanced diet, stay hydrated, and consider taking supplements like vitamin B complex, magnesium, and omega-3 fatty acids, which may help with mood stabilization and reduce anxiety.  7. Emotional support: Ensure that patients have access to counseling services or support groups to address the psychological aspects of opioid withdrawal. Cognitive-behavioral therapy (CBT) can be particularly helpful.  8. Tapering approach: Review the current opioid tapering schedule to ensure it is appropriate for each patient. Sometimes, slowing down the taper or adopting a more gradual approach can minimize withdrawal symptoms.  In summary, consider adding adjuvant medications like gabapentin or baclofen, utilizing non-pharmacological interventions, optimizing nutritional support, providing emotional support, and adjusting the tapering schedule if necessary. These strategies, combined with clonidine and comfort medications \\ \bottomrule
\end{tabular}
\caption{OpenBioLLM-70B responses to role-based queries, corresponding to the role-agnostic query `\textit{How do individuals deal with intense full body cravings during recovery from opiates?}'}
\label{tab:openbio-70b-responses}
\end{table*}

\end{document}